\documentclass[10pt,twocolumn,letterpaper]{article}

\usepackage[pagenumbers]{wacv}
\usepackage[letterpaper,margin=1in,includefoot,footskip=18pt]{geometry}                %
\usepackage{dblfloatfix}
\usepackage{tikz}
\usepackage[utf8]{inputenc}
\usepackage[T1]{fontenc}
\usepackage{amsfonts}
\usepackage{amssymb}
\usepackage{url}
\usepackage{nicefrac}
\usepackage{microtype}
\usepackage{xcolor}
\usepackage{tabularx}
\usepackage{array}
\usepackage{makecell}
\usepackage{subcaption}

\definecolor{wacvblue}{rgb}{0.21,0.49,0.74}
\usepackage[pagebackref,breaklinks,colorlinks,allcolors=wacvblue]{hyperref}
\usepackage[accsupp]{axessibility}
\def\wacvPaperID{1300} %
\def\confName{WACV}
\def\confYear{2027}

\title{DriveOSS: Dataset and Benchmark for 3D Anomaly Segmentation and Open-set Semantic Segmentation in Autonomous Driving}

\author{
Zhiran Yan$^{1}$ \quad
Savankumar Bhanderi$^{1}$ \quad
Sourabh Lolge$^{1}$ \quad
Rui Song$^{2}$ \\
Girdharbhai Paneliya$^{1}$ \quad
Shiva Agrawal$^{1}$ \quad
Gordon Elger$^{1}$ \\[0.5em]
$^{1}$Technische Hochschule Ingolstadt   \quad $^{2}$UCLA
}

\hypersetup{
  pdftitle={DriveOSS: Dataset and Benchmark for 3D Anomaly Segmentation and Open-set Semantic Segmentation in Autonomous Driving},
  pdfauthor={Zhiran Yan, Savankumar Bhanderi, Sourabh Lolge, Rui Song, Girdharbhai Paneliya, Shiva Agrawal, Gordon Elger}
}

\begin{document}
\maketitle
\newcommand{\IEEEcopyrightyear}{2027}

\begingroup
\renewcommand{\thefootnote}{}
\footnotetext{%
\copyright\ \IEEEcopyrightyear\ IEEE. Personal use of this material is permitted.
Permission from IEEE must be obtained for all other uses, in any current or
future media, including reprinting/republishing this material for advertising
or promotional purposes, creating new collective works, for resale or
redistribution to servers or lists, or reuse of any copyrighted component of
this work in other works.%
}
\endgroup

\begin{abstract}
Recent advances in 3D semantic segmentation have significantly improved scene understanding in autonomous driving. However, existing methods are typically developed under a closed-set assumption and evaluated under standard driving conditions, where all categories are known in advance. In real-world driving, autonomous vehicles inevitably encounter previously unseen objects, making closed-set perception insufficient for safe deployment. To address this challenge, we introduce DriveOSS, which, to the best of our knowledge, is the first real-world multimodal dataset and benchmark specifically designed for 3D anomaly segmentation and open-set semantic segmentation. It contains 12,661 synchronized and annotated frames captured using six surround-view RGB cameras, a 128-beam LiDAR, and a GPS/IMU unit. The dataset comprises 52 real-world driving scenes collected under clear, nighttime, cloudy, rainy, and snowy conditions and is organized into ID (in-distribution) and OOD (out-of-distribution) subsets. The ID scenes contain only known classes and are used for training, whereas the OOD scenes additionally contain unknown objects annotated as the outlier class and are used for validation and testing. Extensive experiments show that unseen objects remain a major challenge for current 3D perception systems. The dataset, benchmark protocols, and code are available at \href{https://github.com/chrisyan/DriveOSS}{https://github.com/chrisyan/DriveOSS}.

\end{abstract}
\section{Introduction}
\label{sec:intro}

\begin{figure*}[t]
  \centering
  \includegraphics[width=\textwidth]{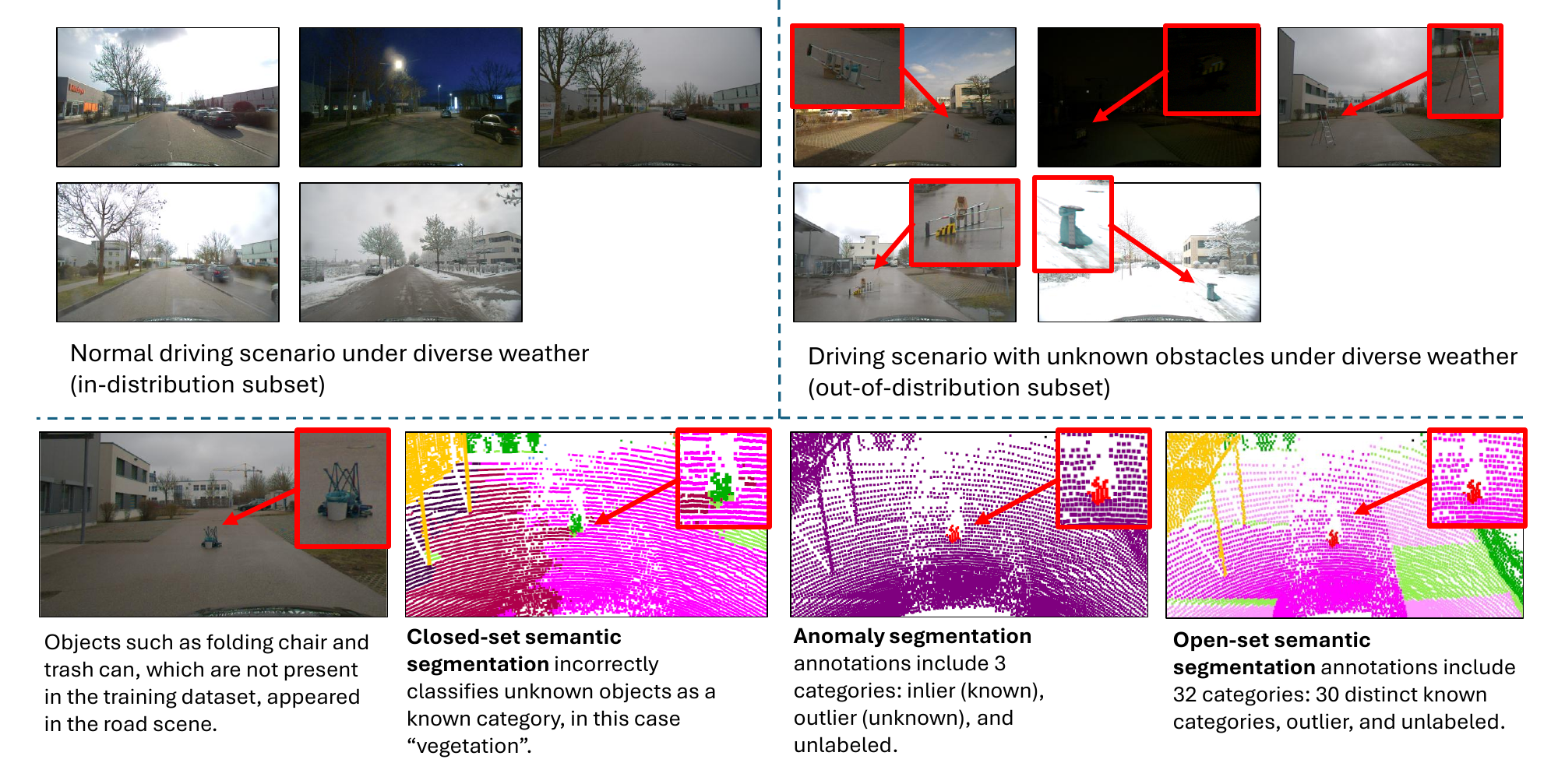}
\caption{ Overview of the proposed DriveOSS dataset and benchmark. \textbf{Top:} Data collected under diverse weather and illumination conditions, including clear, nighttime, cloudy, rainy, and snowy scenes. The left shows the in-distribution (ID) subset containing normal driving scenarios, while the right shows the out-of-distribution (OOD) subset containing unknown objects that are not used during training. \textbf{Bottom:} Different perception settings illustrated using the same scene. From left to right: RGB image, closed-set semantic segmentation prediction, anomaly segmentation ground truth, and open-set semantic segmentation ground truth. }
  \label{figure_1}
\end{figure*}

Robust 3D scene understanding is essential for autonomous driving. Although recent semantic segmentation methods achieve strong performance on established benchmarks, they typically operate under a \emph{closed-set} assumption, assigning every point to a predefined class. This assumption is violated when vehicles encounter unknown obstacles, such as construction materials or fallen cargo. Misclassifying such objects as known classes may lead to unreliable downstream decisions. Open-set semantic segmentation therefore requires jointly recognizing in-distribution (ID) classes and detecting out-of-distribution (OOD) objects.

While open-set segmentation has received increasing attention in 2D vision~\cite{rai2024mask2anomaly,hendrycks2019scaling,grcic2022densehybrid,gasperini2023segmenting}, progress in 3D remains limited. Existing studies generally evaluate closed-set semantic segmentation and anomaly detection separately, although real-world systems require both capabilities. Moreover, standardized 3D open-set benchmarks remain scarce, particularly under adverse weather and illumination~\cite{nekrasov2025spotting,sakaridis2021acdc}. Most driving datasets emphasize favorable daytime conditions and insufficiently represent nighttime, rain, snow, and reduced visibility~\cite{behley2019semantickitti,nekrasov2025spotting}, which complicate both known-class recognition and unknown-object detection.

To address this gap, we introduce \textbf{DriveOSS}, which, to the best of our knowledge, is the first real-world multimodal dataset and benchmark specifically designed for 3D open-set semantic segmentation. It provides synchronized LiDAR point clouds, surround-view RGB images, and point-wise annotations for known classes and unknown objects across diverse locations, weather, and illumination conditions. DriveOSS comprises separate ID and OOD subsets and supports closed-set semantic segmentation, anomaly segmentation, and their joint evaluation through 3D open-set semantic segmentation.

Our contributions are summarized as follows:
\begin{itemize}
    \itemsep0em

    \item We introduce \textbf{DriveOSS}, a real-world multimodal dataset for 3D open-set perception, containing synchronized LiDAR and surround-view RGB data with point-wise annotations under diverse weather and illumination conditions.

    \item We establish benchmarks for 3D anomaly segmentation and 3D open-set semantic segmentation, complemented by closed-set semantic evaluation on the ID subset.

    \item We adapt open-IoU from pixels to 3D points to jointly evaluate known-class recognition and unknown-object detection.

    \item We benchmark representative LiDAR-only and LiDAR--camera fusion models using post-hoc anomaly scores and the feature-based method, revealing substantial gaps between closed-set accuracy, anomaly ranking, and open-set performance.
\end{itemize}

\section{Related Work}
\label{sec:related_works}

\textbf{Autonomous Driving Datasets for 3D Closed-Set Segmentation.}
Large-scale datasets have driven substantial progress in autonomous-driving perception. Early benchmarks such as KITTI~\cite{geiger2013vision}, nuScenes~\cite{caesar2020nuscenes}, and the Waymo Open Dataset~\cite{sun2020scalability} established standard tasks including detection, tracking, and scene understanding. Subsequent datasets, including SemanticKITTI~\cite{behley2019semantickitti}, Toronto-3D~\cite{tan2020toronto3d}, A2D2~\cite{geyer2020a2d2}, SemanticPOSS~\cite{pan2020semanticposs}, PandaSet~\cite{xiao2021pandaset}, Panoptic nuScenes~\cite{fong2022panoptic}, SemanticUSL~\cite{jiang2021lidarnet}, KITTI-360~\cite{liao2022kitti360}, and R2D2~\cite{ninan2023r2d2}, introduced dense annotations for 3D semantic scene understanding. However, these datasets primarily adopt a closed-set taxonomy and are not designed to evaluate unknown-object recognition. Moreover, adverse weather and low-visibility conditions remain underrepresented in many existing benchmarks.

\begin{table*}[ht]
\centering
\small
\caption{\textbf{Comparison of representative automotive 3D segmentation datasets.}
SS, IS, PS, AS, OSS, and SSC denote semantic, instance, panoptic, anomaly, and open-set semantic segmentation, and semantic scene completion, respectively. L and C denote LiDAR and camera, while \#Beam indicates the number of LiDAR beams. The ID Evaluation Classes column reports the number of ID semantic classes used for benchmark evaluation.
$^{a}$SemanticKITTI provides 34 raw labels, which are merged or ignored to obtain 19 ID evaluation classes.
$^{b}$DriveOSS provides 32 raw annotation labels, including \emph{outlier} and \emph{unlabeled}. After merging semantically equivalent labels and excluding ignored labels, 19 ID semantic classes are retained for training and evaluation. The \emph{outlier} label is additionally used for anomaly segmentation and open-set semantic segmentation.}
\label{tab:dataset_comparison}

\resizebox{\textwidth}{!}{%
\begin{tabular}{l c c c c c c c c c}
\toprule
Dataset
& Year
& Weather
& Modality
& \#Beam
& Camera
& \#Scan
& ID Evaluation Classes
& OOD Scene Labels
& Supported 3D Tasks \\
\midrule

SemanticKITTI~\cite{behley2019semantickitti}
& 2019
& Clear
& L+C
& 64
& Front-view
& 43,552
& 19$^{a}$
& --
& SS, PS \\

Waymo~\cite{sun2020scalability}
& 2019
& Clear/Night/Rainy
& L+C
& 64
& Surround-view
& 46,000
& 23
& --
& SS \\

Toronto-3D~\cite{tan2020toronto3d}
& 2020
& Clear
& L+C
& 32
& Surround-view
& --
& 8
& --
& SS \\

A2D2~\cite{geyer2020a2d2}
& 2020
& Cloudy/Rainy/Sunny
& L+C
& 16
& Surround-view
& 41,277
& 38
& --
& 2D SS, detection \\

SemanticPOSS~\cite{pan2020semanticposs}
& 2020
& Clear
& L+C
& 40
& Surround-view
& 2,988
& 14
& --
& SS, PS \\

PandaSet~\cite{xiao2021pandaset}
& 2021
& Clear/Night
& L+C
& 64/150
& Surround-view
& 16,480
& 37
& --
& SS, detection \\

Panoptic nuScenes~\cite{fong2022panoptic}
& 2021
& Diverse
& L+C
& 32
& Surround-view
& 40,000
& 16
& --
& PS \\

SemanticUSL~\cite{jiang2021lidarnet}
& 2021
& Clear
& L
& 64
& --
& 1,200
& 20
& --
& SS \\

KITTI-360~\cite{liao2022kitti360}
& 2022
& Clear
& L+C
& 64
& Surround-view
& 100k
& 37
& --
& SS, IS, SLAM, SSC \\

Navya3DSeg~\cite{10168231}
& 2023
& Night
& L
& 40
& --
& 50k
& 30
& --
& SS \\

R2D2~\cite{ninan2023r2d2}
& 2023
& Clear
& L+C
& 128
& Front-view
& 10.5k
& 28
& --
& SS \\

SemanticSTF~\cite{xiao20233d}
& 2023
& Fog/Rain/Snow
& L+C
& 64
& Front-view
& 2,076
& 20
& --
& SS \\

WildScenes~\cite{vidanapathirana2025wildscenes}
& 2024
& Diverse
& L+C
& 16
& Front-view
& 12,148
& 15
& --
& SS \\

SemanticTHAB~\cite{reichert2025real}
& 2025
& Clear
& L
& 128
& --
& 4,750
& 20
& --
& SS \\

Forklifts~\cite{serfling2025lidar}
& 2025
& Clear
& L
& 128
& --
& 7,676
& 9
& --
& SS \\

Panoptic-CUDAL~\cite{11423256}
& 2025
& Rain
& L+C
& 128
& Surround-view
& 14.7k
& 14
& --
& PS \\

STU~\cite{nekrasov2025spotting}
& 2025
& Clear
& L+C
& 128
& Surround-view
& 9,982
& 27
& Inlier + outlier + unlabeled
& PS, AS \\

LIDO~\cite{Mosco_2026_CVPR}
& 2026
& Clear
& L
& 32/40/64
& --
& 10,590
& 13/16/19
& Inlier + outlier + unlabeled
& AS \\

\midrule

\textbf{DriveOSS (Ours)}
& \textbf{2026}
& \makecell{\textbf{Clear/Night/}\\
            \textbf{Cloudy/Rain/Snow}}
& \textbf{L+C}
& \textbf{128}
& \textbf{Surround-view}
& \textbf{12,661}
& \textbf{19}$^{b}$
& \makecell{\textbf{19 ID classes +}\\
            \textbf{outlier + unlabeled}}
& \textbf{SS, AS, OSS} \\

\bottomrule
\end{tabular}%
}
\end{table*}

\textbf{Autonomous Driving Datasets for Open-Set Perception.}
Recognizing unseen objects is essential for the safe deployment of autonomous systems. In 2D vision, benchmarks such as Lost and Found~\cite{pinggera2016lost}, Fishyscapes~\cite{blum2021fishyscapes}, RoadAnomaly~\cite{lis2019detecting}, SegmentMeIfYouCan~\cite{chan2021segmentmeifyoucan}, StreetHazards~\cite{hendrycks2019scaling}, SOD~\cite{singh2020lidar}, SOS~\cite{maag2022sos}, and CODA~\cite{li2022coda} have advanced anomaly detection and open-set scene understanding.

Extending open-set perception to 3D is more challenging because LiDAR measurements are sparse and affected by range, beam patterns, occlusion, and sensor-specific characteristics. Synthetic benchmarks such as AnoVox~\cite{bogdoll2024anovox} and SDAC~\cite{gong2024sdac} provide scalable evaluation using CARLA~\cite{dosovitskiy2017carla}, but retain a domain gap from real sensor measurements.
LIDO~\cite{Mosco_2026_CVPR} constructs mixed real--synthetic OOD benchmarks from SemanticKITTI, SemanticPOSS, and nuScenes by inserting ModelNet objects into real LiDAR scans. Its generation pipeline adapts the inserted points and intensities to the corresponding LiDAR sensors. Nevertheless, synthetic insertion cannot fully reproduce the material-dependent reflections, occlusions, noise, and sampling patterns of physically measured objects. STU~\cite{nekrasov2025spotting} is among the first real-world multimodal datasets for LiDAR anomaly segmentation, but is limited to clear daytime conditions, collapses OOD scene labels into binary inlier/outlier categories, and evaluates anomaly detection separately from semantic segmentation rather than in a unified open-set setting.

DriveOSS addresses these limitations by capturing real OOD obstacles across clear, nighttime, cloudy, rainy, and snowy conditions while retaining full 19-class ID annotations in OOD scenes to benchmark closed-set, anomaly, and joint 3D open-set semantic segmentation across LiDAR and camera-LiDAR models.

\section{DriveOSS Dataset}
\label{sec:dataset}

\subsection{Data Acquisition}
\label{data_acquisition}

\begin{figure}[t]
    \centering
    \includegraphics[width=0.68\columnwidth]{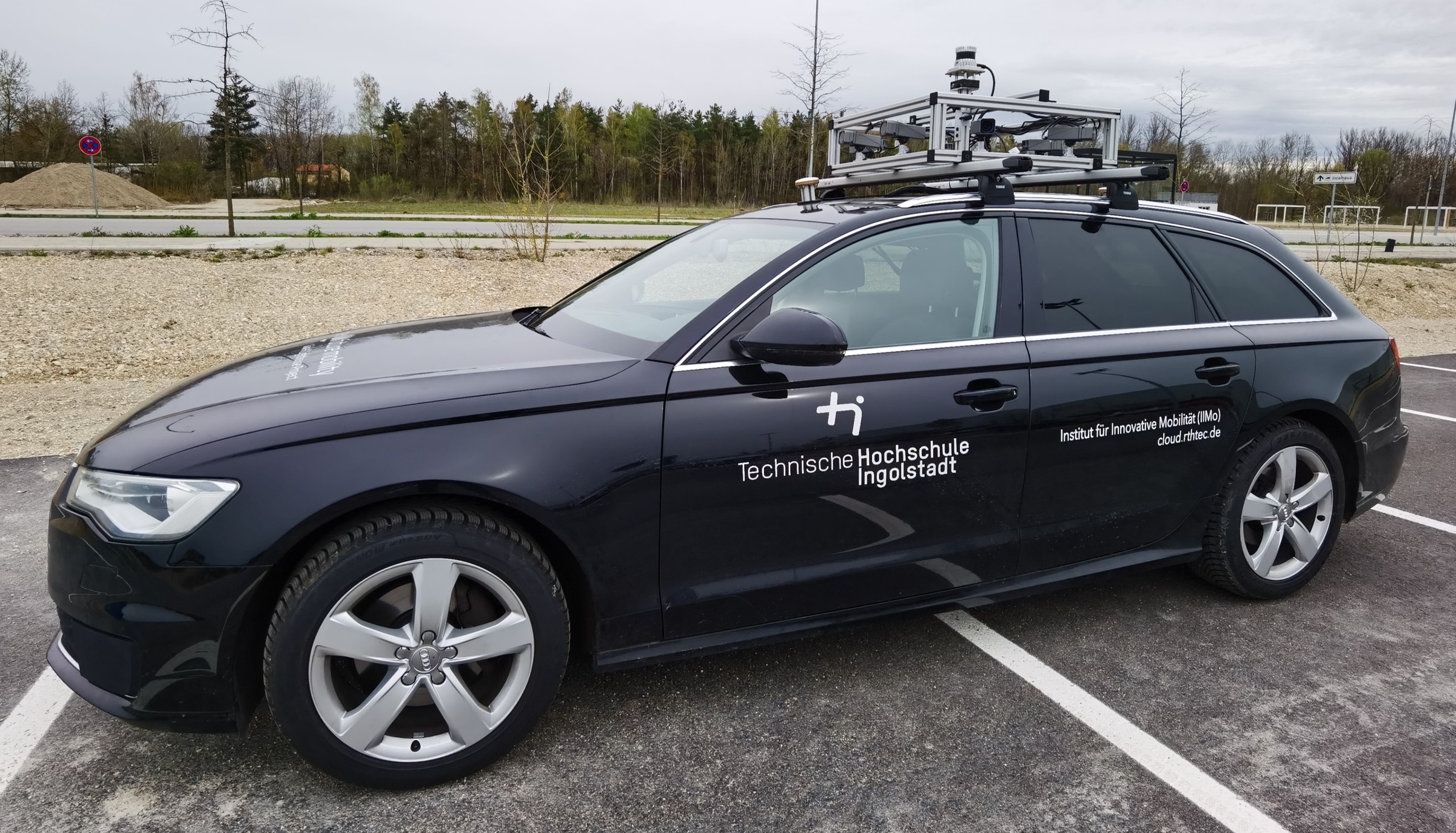}

    \vspace{1mm}

    \includegraphics[width=0.80\columnwidth]{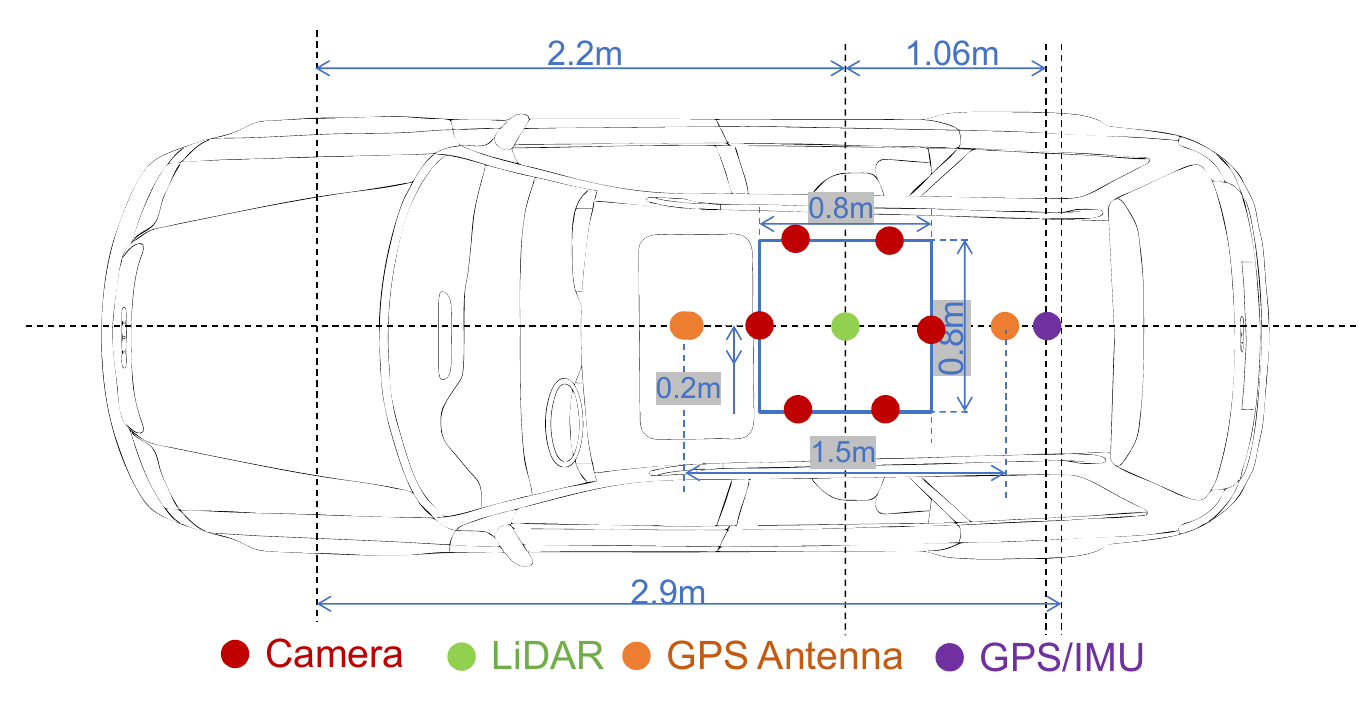}

    \caption{\textbf{DriveOSS data collection platform.}
    Top: the sensor vehicle. Bottom: six surround-view cameras, one
    360$^\circ$ LiDAR, and one GPS/IMU unit.}
    \label{fig:sensor_layout}
    \vspace{-5mm}
\end{figure}

DriveOSS was collected using a vehicle equipped with an Ouster OS1-128 LiDAR, six Basler RGB cameras, and a Certus GPS/IMU system (Fig.~\ref{fig:sensor_layout}). All sensors are synchronized to a common GPS-derived time reference. The cameras operate at 10\,Hz, the LiDAR at 20\,Hz, and the GPS/IMU at 600\,Hz. Detailed hardware specifications are provided in the supplementary material.

\begin{table*}[!b]
  \centering
  \caption{\textbf{Taxonomy of OOD objects in DriveOSS.}}
  \label{tab:ood_taxonomy}
  \small
  \begin{tabularx}{\textwidth}{>{\raggedright\arraybackslash}p{0.23\textwidth} >{\raggedright\arraybackslash}X >{\raggedright\arraybackslash}X}
    \toprule
    Category & Representative Objects & Characteristics \\
    \midrule

    Containers \& Packaging
    & Cardboard boxes, plastic boxes, foam blocks, garbage bags, and small trash bins
    & Common roadside obstacles with simple geometries and diverse materials \\

    Furniture \& Fixtures
    & Office chairs, folding chairs, wooden chairs, clothes racks, and advertising signs
    & Bulky objects with complex structures and varying heights \\

    Tools \& Industrial Equipment
    & Ladders, shovels, metal barriers, and maintenance equipment
    & Rigid structures with distinctive geometries and occasionally high reflectivity \\

    Personal Belongings
    & Backpacks, bicycle helmets, bags, and other personal items
    & Predominantly small objects with irregular shapes and mixed materials \\

    Utility \& Service Equipment
    & Shopping carts, waste-collection carts, and mobile bins
    & Movable service-related objects commonly encountered in urban environments \\

    Roadside Debris \& Miscellaneous
    & Wooden boards, fallen signs, and other uncommon objects
    & Unstructured obstacles with diverse appearances and physical properties \\

    \bottomrule
  \end{tabularx}
\end{table*}

\begin{figure}[h]
  \centering
  \includegraphics[width=\columnwidth]{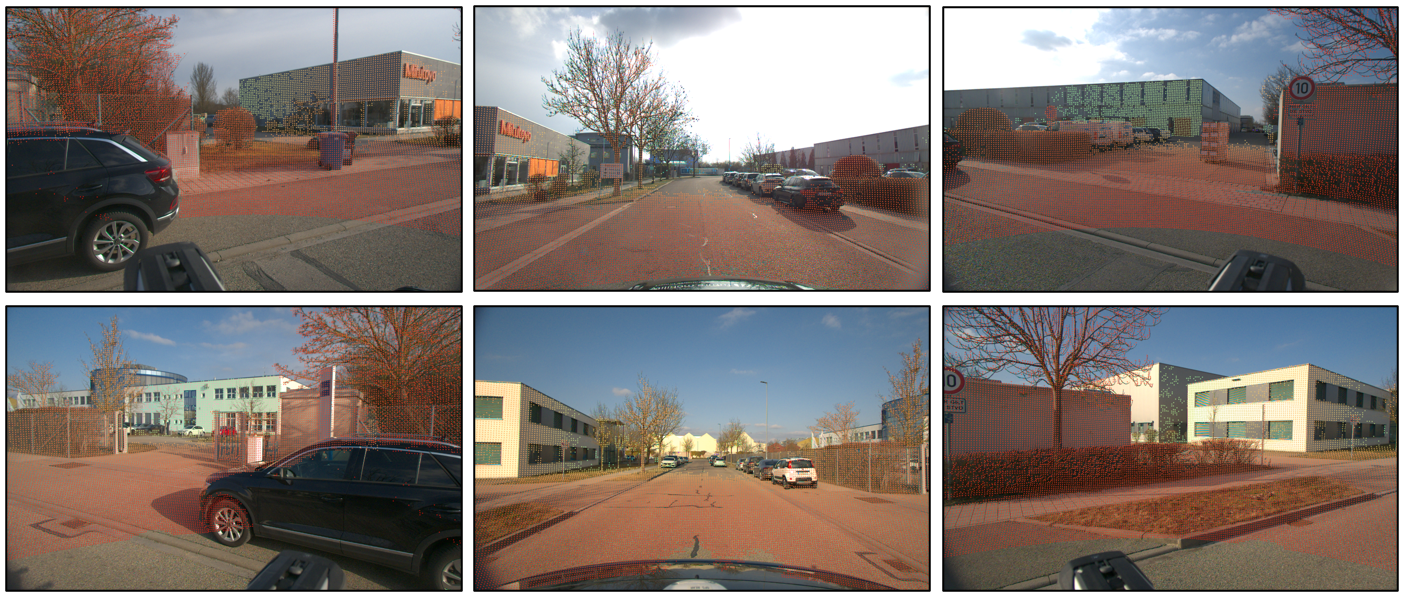}
  \caption{\textbf{LiDAR--camera calibration.}
  LiDAR points projected onto the six camera views using the estimated
  intrinsic and extrinsic parameters.}
  \label{fig:calibration}
  \vspace{-4mm}
\end{figure}

Camera intrinsics are estimated using Zhang's method~\cite{zhang2000flexible}, while LiDAR-to-camera extrinsics are obtained from manually selected 2D--3D correspondences using PnP~\cite{lepetit2009epnp}. Figure~\ref{fig:calibration} visualizes the resulting LiDAR projections onto all six camera views.

\begin{figure}[h]
  \centering
  \includegraphics[width=\linewidth]{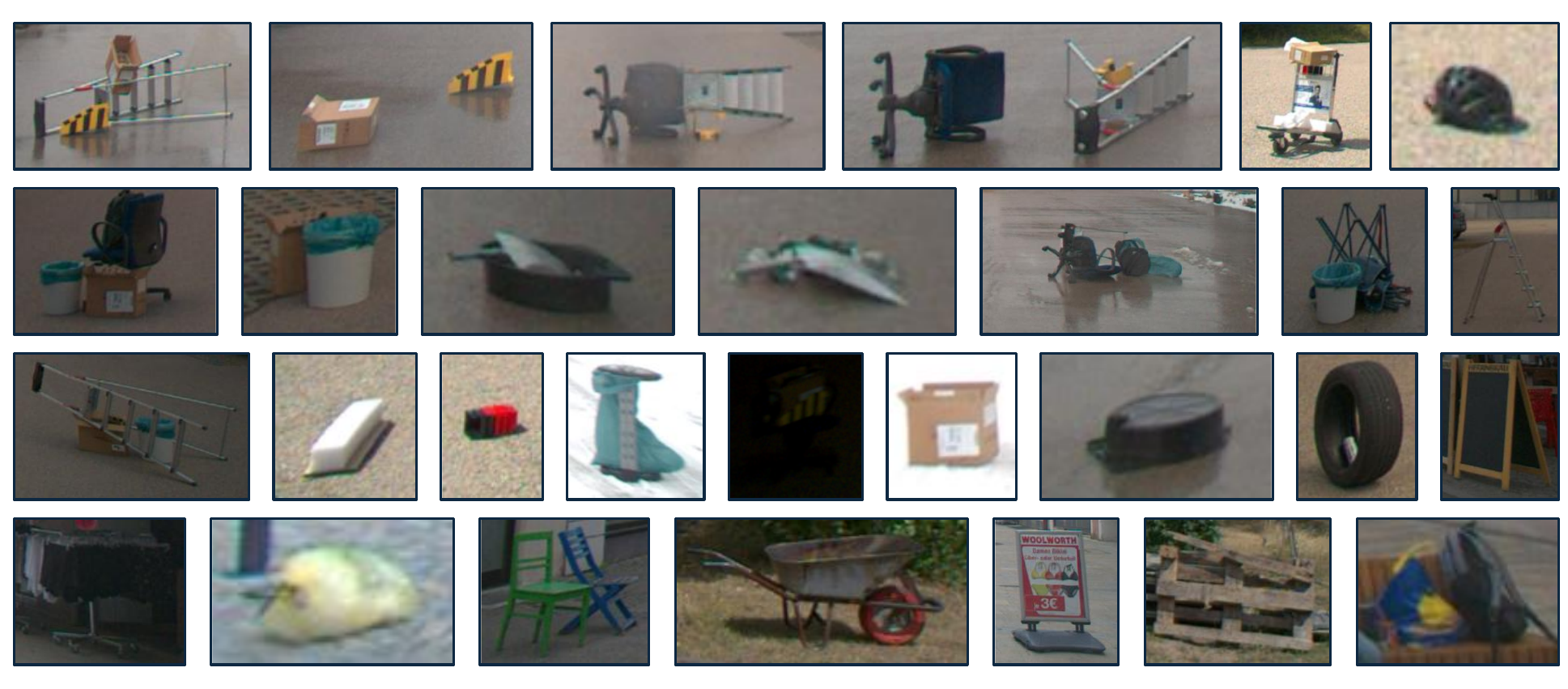}
  \caption{\textbf{Representative anomalies in DriveOSS.}
  Physically present objects with diverse scales, geometries, and
  materials, captured under different environmental conditions.}
  \label{fig:anomaly_objects}
  \vspace{-4mm}
\end{figure}

Data were recorded in downtown areas, pedestrian zones, urban and rural roads, parking lots, and industrial areas under clear, cloudy, rainy, snowy, daytime, and nighttime conditions. DriveOSS-ID contains normal driving scenes without benchmark outliers. DriveOSS-OOD contains both controlled anomalies, safely placed on or alongside the road, and naturally occurring anomalies encountered during regular driving (Fig.~\ref{fig:anomaly_objects}). Before release, faces and license plates are anonymized through automatic blurring followed by manual verification.

The anomalous objects are organized into six functional categories (Table~\ref{tab:ood_taxonomy}) and span diverse scales, geometries, and materials. Although plausible in driving environments, they fall outside the evaluated ID taxonomy and are designated as benchmark outliers.

\subsection{Data Annotation}
\label{data_annotation}
Following STU~\cite{nekrasov2025spotting}, we define anomalies as unknown physical obstacles that pose potential collision or navigation risks to autonomous vehicles. To ensure an objective boundary between \emph{outlier} and \emph{other-object}, we enforce a strict spatial-hazard protocol: any object outside the 19 evaluated ID classes located on or immediately adjacent to the drivable surface (road, parking, lane-marking, or sidewalk curb) is annotated as \emph{outlier}. We follow the same taxonomy for \emph{other-object} class as SemanticKITTI. Furthermore, objects outside the ID taxonomy that are permanent off-road background fixtures (e.g., bus shelters, benches, or far-sidewalk bins) are labeled as \emph{other-object}. Noisy returns and points outside the region of interest are labeled as \emph{unlabeled}. The \emph{other-object}, \emph{other-structure}, and \emph{unlabeled} categories are excluded from both ID and OOD benchmark evaluation.

\begin{figure*}[t]
  \centering
  \includegraphics[width=\textwidth]
  {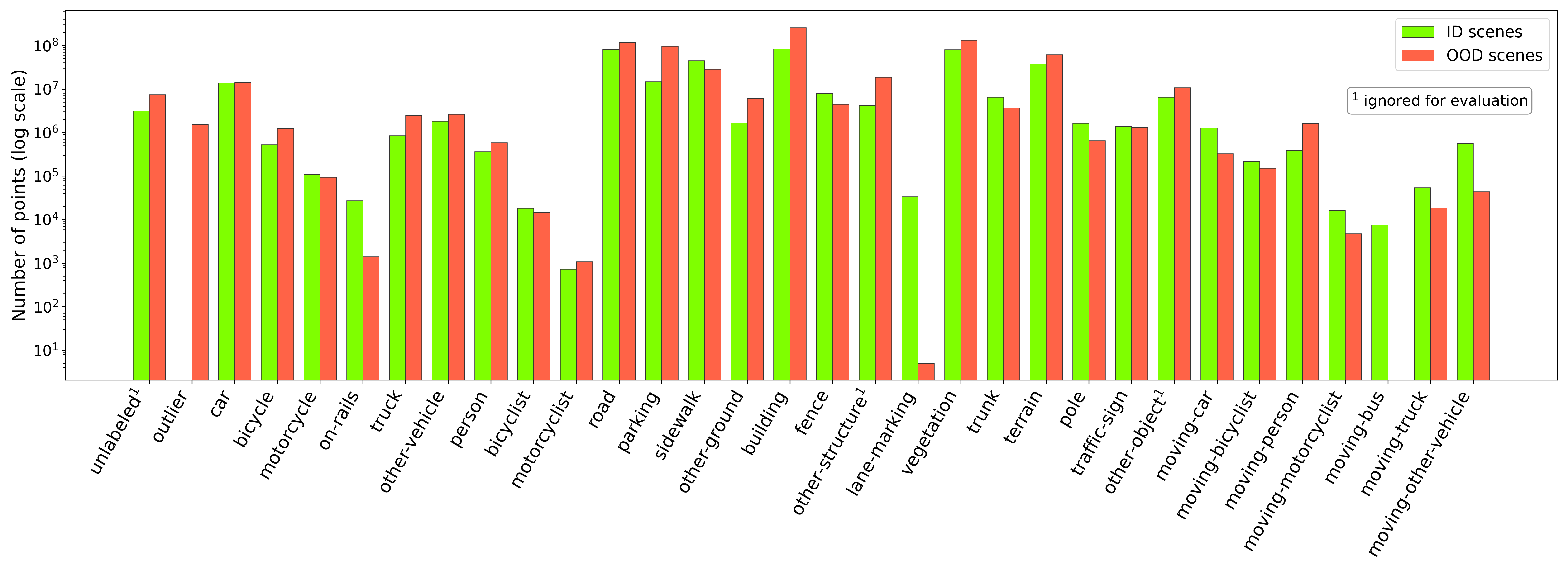}
  \caption{\textbf{Label distributions in DriveOSS-ID and DriveOSS-OOD.}
  The \emph{outlier} label occurs exclusively in DriveOSS-OOD, while
  labels marked with $^{1}$ are excluded from evaluation.}
  \label{fig:label_distribution}
  \vspace{-3mm}
\end{figure*}

Unlike anomaly-only datasets that reduce scenes to inlier, outlier, and unlabeled, DriveOSS retains fine-grained semantic annotations for known classes. The raw annotation space comprises 32 labels (Fig. \ref{fig:label_distribution}). After merging dynamic counterparts into their static semantic classes (e.g., \emph{moving-car} to \emph{car}) and excluding ignored labels, 19 ID classes aligned with SemanticKITTI are retained alongside the 20th \emph{outlier} class.

For efficient and robust ground truth generation, our annotation pipeline uses automatic pre-labeling strictly as a coarse initial seed for known classes, followed by extensive manual refinement. Specifically, Mask4Former~\cite{yilmaz2024mask4former} pretrained on SemanticKITTI generates initial closed-set predictions, and KISS-ICP~\cite{vizzo2023kiss} registers consecutive scans for loading into the SemanticKITTI Point Labeler ~\cite{behley2019semantickitti}. Annotators then manually refine object and ground boundaries using ground/plane removal and class-isolation filters to correct misclassified points. Surround-view images are used separately by the annotators for better understanding of the environment. All annotators strictly adhere to a unified labeling policy, and every sequence is independently cross-checked and verified by at least one additional annotator. Visual comparisons between pre-labels and final ground truth are provided in the supplementary material.

\begin{table*}[!b]
\centering
\small
\caption{\textbf{Distribution across environmental conditions.}
Numbers of sequences and LiDAR frames for each split, together with
annotated OOD points for DriveOSS-OOD.}
\label{tab:driveoss_condition_statistics}
\resizebox{0.9\textwidth}{!}{
\begin{tabular}{lrrrrrrrrrr}
\toprule
& \multicolumn{2}{c}{ID Train}
& \multicolumn{2}{c}{ID Validation}
& \multicolumn{3}{c}{OOD Validation}
& \multicolumn{3}{c}{OOD Test} \\
\cmidrule(lr){2-3}
\cmidrule(lr){4-5}
\cmidrule(lr){6-8}
\cmidrule(lr){9-11}
Condition
& Seq. & Frames
& Seq. & Frames
& Seq. & Frames & OOD Points
& Seq. & Frames & OOD Points \\
\midrule
Clear
& 4 &   809
& 6 & 1,701
& 9 & 2,315 & 252,606
& 8 & 1,766 & 885,729 \\

Nighttime
& 1 &   184
& 1 &    79
& 2 &   287 & 22,132
& 2 &   357 & 42,979 \\

Cloudy
& 2 &   636
& 1 &   186
& 2 &   742 & 43,548
& 2 &   666 & 71,440 \\

Rainy
& 1 &   399
& 1 &   165
& 2 &   696 & 35,104
& 2 &   618 & 66,509 \\

Snowy
& 1 &   212
& 1 &   253
& 2 &   238 & 12,898
& 2 &   352 & 89,803 \\

\midrule
Overall
& 9 & 2,240
& 10 & 2,384
& 17 & 4,278 & 366,288
& 16 & 3,759 & 1,156,460 \\
\bottomrule
\end{tabular}}
\vspace{-3mm}
\end{table*}

\subsection{Data Analysis}
\label{data_analysis}

\begin{figure}[ht]
  \centering
  \includegraphics[width=0.85\columnwidth]
  {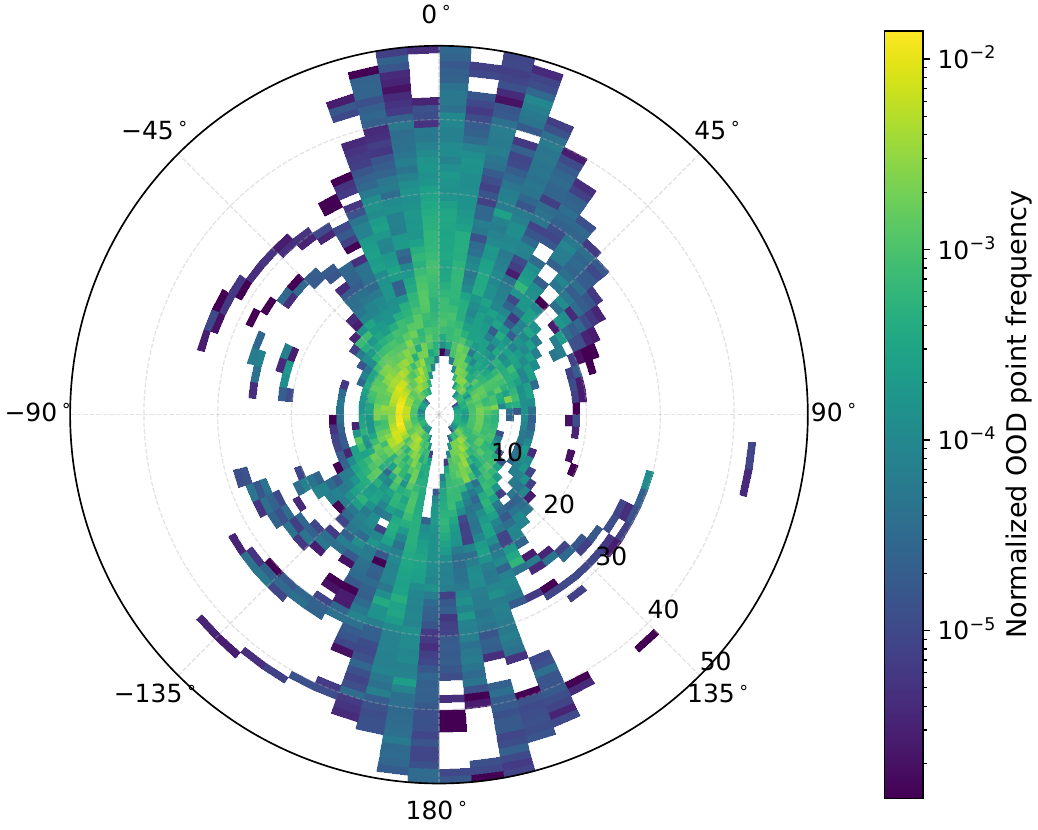}
  \caption{\textbf{Spatial distribution of OOD points.}
  Normalized OOD-point frequency relative to the ego vehicle over all
  DriveOSS-OOD validation and test sequences.}
  \label{fig:ood_spatial_distribution}
  \vspace{-3mm}
\end{figure}

DriveOSS contains 19 ID and 33 OOD sequences recorded at geographically disjoint locations. The acquisition routes of the two subsets do not overlap, reducing the possibility of exploiting location-specific background cues and enabling evaluation in unseen environments.
The data are sampled at 10\,Hz. Using the cameras as the reference modality, each synchronized six-view image set is associated with the temporally closest LiDAR and GPS/IMU measurements. In total, DriveOSS provides 12,661 LiDAR frames, 75,966 camera images, and 12,661 synchronized GPS/IMU measurements. DriveOSS-ID is used for closed-set training and validation, whereas DriveOSS-OOD is divided into validation and test sets for anomaly and open-set semantic segmentation.
Figure~\ref{fig:label_distribution} shows the point-level label distributions of both subsets. They share the same 19-class ID mapping, while \emph{outlier} occurs exclusively in DriveOSS-OOD. Table~\ref{tab:driveoss_condition_statistics} reports the distribution across the four benchmark splits and five environmental conditions. DriveOSS contains 9 ID training, 10 ID validation, 17 OOD validation, and 16 OOD test sequences, comprising 12,661 LiDAR frames and 1,522,748 annotated OOD points. Figure~\ref{fig:ood_spatial_distribution} further visualizes the spatial coverage of OOD points relative to the ego vehicle.

\section{Benchmark Tasks}
\label{tasks}

\subsection{Closed-Set 3D Semantic Segmentation}
\label{tasks_1}

\textbf{Task Definition.}
Given a LiDAR point cloud $\mathcal{X}=\{p_i\}_{i=1}^{N}$, optionally accompanied by synchronized RGB images, the objective is to assign each point $p_i$ a semantic label from the predefined set of ID classes.
\textbf{Evaluation Metrics.}
Closed-set semantic segmentation is evaluated on the 19 ID classes using class-wise intersection over union (IoU) and mean IoU (mIoU).
\textbf{Benchmark Methods.}
We evaluate the LiDAR-only models Mask4Former3D~\cite{nekrasov2025spotting}, P3Former~\cite{xiao2025position}, LCPS-lidar~\cite{Zhang_2023_ICCV}, and LIDO~\cite{Mosco_2026_CVPR}, together with the LiDAR--camera fusion model LCPS. Although LIDO is primarily designed for anomaly segmentation, its semantic segmentation branch also produces closed-set predictions, which are evaluated using the same 19-class protocol. We additionally extend Mask4Former3D with a camera branch and query-level cross-attention to construct a multimodal fusion baseline, denoted as Mask4Former3D*. Additional details are provided in the supplementary material.

\subsection{3D Anomaly Segmentation}
\label{tasks_2}

\textbf{Task Definition.}
Given a LiDAR point cloud $\mathcal{X}=\{p_i\}_{i=1}^{N}$, optionally accompanied by synchronized RGB images, the model predicts a continuous anomaly score for each point. A higher score indicates a greater likelihood that the point belongs to an OOD object.
\textbf{Evaluation Metrics.}
Anomaly segmentation is formulated as binary discrimination between ID and OOD points, with ignored labels excluded from evaluation. Because the evaluated methods produce continuous anomaly scores, we report the area under the receiver operating characteristic curve (AUROC), average precision (AP), and the false positive rate at 95\% true positive rate (FPR@95)~\cite{blum2021fishyscapes}. AUROC and AP evaluate anomaly-score ranking across all possible thresholds, whereas FPR@95 measures the proportion of ID points incorrectly classified as OOD when 95\% of OOD points are detected.
\textbf{Benchmark Methods.}
We evaluate two categories of anomaly segmentation methods. First, we apply representative post-hoc scoring functions to the outputs of the closed-set models described in Sec.~\ref{tasks_1}, including maximum softmax probability (MSP)~\cite{hendrycks2016baseline}, MaxLogit~\cite{hendrycks2019scaling}, entropy~\cite{chan2021entropy}, energy~\cite{tian2022pixel}, and RbA~\cite{nayal2023rba}. Second, we benchmark LIDO~\cite{Mosco_2026_CVPR}, which jointly produces semantic predictions and feature-based anomaly scores. All methods are trained exclusively on ID data without using DriveOSS-OOD samples or anomaly annotations.

\subsection{3D Open-Set Semantic Segmentation}
\label{tasks_3}

\textbf{Task Definition.}
Open-set semantic segmentation jointly addresses known-class recognition and unknown-object detection. Each point is assigned either to one of the 19 ID semantic classes or to an additional \emph{outlier} class.
\textbf{Evaluation Metrics.}
Following image-domain open-set semantic segmentation protocols such as DenseHybrid~\cite{grcic2022densehybrid}, we extend open-IoU evaluation to the 3D point-cloud domain. Open-set predictions are constructed by combining closed-set semantic predictions with point-wise anomaly scores. Points with anomaly scores exceeding a fixed threshold are reassigned to the \emph{outlier} class, while the remaining points retain their closed-set semantic predictions. The threshold is calibrated on the DriveOSS-OOD validation set and fixed for test-set evaluation. We report class-wise open-IoU and mean open-IoU (mOpen-IoU) over the 19 ID classes, together with the IoU of the outlier class. These metrics jointly capture known-class segmentation quality and errors introduced by unknown-object detection. Further details and exact mathematical formulation of open-IoU are provided in the supplementary material.
\textbf{Benchmark Methods.}
For the post-hoc baselines, open-set predictions are constructed by combining the semantic predictions of the closed-set models from Sec.~\ref{tasks_1} with their corresponding anomaly scores described in Sec.~\ref{tasks_2}. For LIDO, its own semantic predictions and feature-based anomaly scores are combined to produce the final open-set predictions. All methods are evaluated using the same validation-based threshold calibration protocol.

\section{Experiments}
\label{experiments}

\subsection{Implementation Details}

All trainable models are trained exclusively using ID semantic classes. To increase training-data diversity and reduce dataset- and sensor-specific domain gaps, we augment DriveOSS-ID with the training sets of SemanticKITTI~\cite{behley2019semantickitti} and Panoptic-CUDAL~\cite{11423256}. Panoptic-CUDAL was captured using the same Ouster OS1-128 LiDAR model as DriveOSS and therefore provides complementary training data with similar sensor characteristics. Before joint training, the annotations of all three datasets are mapped to a unified 19-class label space. Neither DriveOSS-OOD samples nor anomaly annotations are used during training.
For the post-hoc baselines, anomaly scores are derived from the outputs of the trained closed-set segmentation models. LIDO is trained using the same ID-only data and jointly produces closed-set semantic predictions and feature-based anomaly scores.
For open-set evaluation, the operating threshold $\tau_{95}$ is calibrated separately for each method on the DriveOSS-OOD validation set by selecting the anomaly-score threshold corresponding to a 95\% OOD true positive rate. The resulting threshold is fixed and applied to the DriveOSS-OOD test set without further adjustment. Points with anomaly scores above $\tau_{95}$ are assigned to the \emph{outlier} class, whereas the remaining points retain their corresponding closed-set semantic predictions. Further architectural, training, and evaluation details are provided in the supplementary material.

\subsection{Benchmark Results}

\paragraph{Closed-set 3D Semantic Segmentation.}
Table~\ref{tab:closed_semantic_summary} summarizes closed-set semantic segmentation performance on the DriveOSS-ID validation set and the ID-labeled points of the DriveOSS-OOD test set. LIDO achieves the highest mIoU on both splits, reaching 58.35\% and 40.99\%, respectively, and outperforms the second-best method on DriveOSS-OOD by 6.99 points. Complete class-wise results are provided in the supplementary material.
All methods degrade on DriveOSS-OOD, indicating that this split challenges both unknown-object detection and known-class recognition under environmental and appearance shifts. The two LiDAR--camera pairs further show that fusion gains do not consistently transfer to open-world scenes. LCPS improves over LCPS-lidar by 2.11 points on DriveOSS-ID but only 0.25 points on DriveOSS-OOD. Similarly, M4F3D* improves over M4F3D by 13.46 points on DriveOSS-ID, yet underperforms it by 2.65 points on DriveOSS-OOD. Thus, camera features substantially improve in-domain recognition but may introduce additional sensitivity to distribution shifts.

\begin{table}[t]
\centering
\small
\caption{\textbf{Closed-set 3D semantic segmentation.}
mIoU (\%) on DriveOSS-ID validation and the ID-labeled points of
DriveOSS-OOD test. $\Delta$ denotes the change from ID to OOD.}
\label{tab:closed_semantic_summary}
\resizebox{0.8\columnwidth}{!}{
\begin{tabular}{lcccc}
\toprule
Method & M & ID-val & OOD-test & $\Delta$ \\
\midrule
P3Former   & L   & 43.27 & 29.58 & $-13.69$ \\
LCPS-lidar & L   & 51.94 & 33.75 & $-18.19$ \\
M4F3D      & L   & 34.98 & 31.11 & $-3.87$ \\
LIDO       & L   & \textbf{58.35} & \textbf{40.99} & $-17.36$ \\
\midrule
LCPS       & L+C & 54.05 & 34.00 & $-20.05$ \\
M4F3D*     & L+C & 48.44 & 28.46 & $-19.98$ \\
\bottomrule
\end{tabular}}
\vspace{-4mm}
\end{table}

\paragraph{3D Anomaly Segmentation.}
Table~\ref{tab:anomaly_test} reveals a substantial gap between anomaly ranking and high-recall detection. LIDO achieves the best AUROC of 91.70\% and the lowest FPR@95 of 32.66\%, providing the most balanced performance. In contrast, M4F3D+RbA obtains the highest AP of 12.57\% but an FPR@95 of 100\%, indicating that its high AP does not translate into a usable high-recall operating point.
Anomaly scoring is strongly coupled with the underlying representation. RbA achieves 89.80\% AUROC with P3Former but only 50.65\% with LCPS-lidar and 45.43\% with M4F3D*. Likewise, MSP ranges from 26.35\% AUROC with P3Former to 79.43\% with M4F3D. No post-hoc score therefore dominates across architectures.
Fusion also has an architecture- and score-dependent effect. LCPS improves MaxLogit and energy scoring over LCPS-lidar but not MSP or entropy. In contrast, M4F3D* reduces AUROC and AP for all five scores, with AUROC drops of 23.50 and 22.48 points for energy and RbA. Hence, representations optimized for multimodal known-class recognition do not necessarily preserve the separation required for OOD detection.

\begin{table}[t]
\centering
\small
\caption{\textbf{3D anomaly segmentation on DriveOSS-OOD test.}
We report AUROC~$\uparrow$, FPR@95~$\downarrow$, and AP~$\uparrow$.
ML denotes MaxLogit, and M denotes the input modality. M4F3D* is the
modified LiDAR--camera fusion model.}
\label{tab:anomaly_test}
\resizebox{0.9\columnwidth}{!}{
\begin{tabular}{l c c c c}
\toprule
Method & M & AUROC~$\uparrow$ & FPR@95~$\downarrow$ & AP~$\uparrow$ \\
\midrule

P3Former + MSP     & L & 26.35 & 99.50 & 0.27  \\
P3Former + ML      & L & 34.60 &99.80  &0.35  \\
P3Former + Energy  & L & 37.94 & 99.79 & 0.65 \\
P3Former + Entropy & L & 26.63 & 99.57 & 0.37 \\
P3Former + RbA  & L & 89.80 & 45.94 & 3.53 \\

\midrule
LCPS-lidar + MSP         & L & 78.76 & 47.32 &1.11  \\
LCPS-lidar + ML          & L & 46.06 & 83.09 & 0.37 \\
LCPS-lidar + Energy      & L & 46.92 & 82.59 & 0.37 \\
LCPS-lidar + Entropy     & L & 79.76 & 46.95 & 1.38 \\
LCPS-lidar + RbA     & L & 50.65 & 96.24 & 0.37 \\

\midrule
M4F3D + MSP        & L & 79.43 & 98.74 & 10.27 \\
M4F3D + ML         & L & 76.44 & 97.63 & 10.41 \\
M4F3D + Energy     & L  & 70.77 & 98.15 & 10.65 \\
M4F3D + Entropy    & L & 78.38 & 98.74 & 10.97 \\
M4F3D + RbA    & L & 67.91 & 100.00 & 12.57 \\
\midrule
LIDO     & L & 91.70 & 32.66 & 4.44 \\

\midrule
LCPS + MSP         & L+C & 77.84 & 54.66 & 1.19 \\
LCPS + ML          & L+C & 74.57 & 77.01  & 1.48 \\
LCPS + Energy      & L+C & 74.94 & 77.59 & 1.61 \\
LCPS + Entropy     & L+C & 78.67 & 54.44 & 1.59 \\
LCPS + RbA     & L+C & 61.74 & 89.28 & 1.75 \\

\midrule
M4F3D* + MSP       & L+C  & 70.38 & 96.85 & 7.11 \\
M4F3D* + ML        & L+C & 60.37 & 98.54 & 5.88 \\
M4F3D* + Energy    & L+C  & 47.27 & 99.11 & 3.78 \\
M4F3D* + Entropy   & L+C  & 66.85 & 97.08 & 7.07 \\
M4F3D* + RbA   & L+C  & 45.43 & 100.00 & 5.79 \\

\bottomrule
\end{tabular}}
\vspace{-4mm}
\end{table}

\paragraph{3D Open-set Semantic Segmentation.}
Table~\ref{tab:openset_classwise_test} shows that neither closed-set accuracy nor threshold-independent anomaly ranking reliably predicts open-set performance. LIDO achieves the highest mOpen-IoU of 22.98\% and outlier IoU of 1.07\%, consistent with its lower FPR@95. Nevertheless, its mOpen-IoU remains 18.01 points below its closed-set mIoU on the same test scenes, and its outlier IoU remains low.
M4F3D further illustrates this discrepancy. Despite AP values above 10\%, its FPR@95 ranges from 97.63\% to 100\%, while its best mOpen-IoU under the validation-calibrated threshold is 9.58\%. M4F3D* also has high FPR@95 values, ranging from 96.85\% to 100\%, and achieves a best mOpen-IoU of 11.27\%. False rejection can reduce known-class open-IoU: when all correctly predicted points of an ID class are reassigned to the \emph{outlier} class, no true positives remain and its open-IoU becomes zero. The test-set FPR@95 operating point and the fixed validation-calibrated threshold are selected separately and need not coincide.
Interestingly, M4F3D* improves mOpen-IoU for MSP, entropy, and RbA despite lower AUROC and AP. These gains in known-class mOpen-IoU are not accompanied by improved outlier IoU. For example, M4F3D*+RbA increases mOpen-IoU from 9.58\% to 11.27\%, while outlier IoU decreases from 0.32\% to 0.21\%. This discrepancy motivates reporting outlier IoU separately from mean known-class open-IoU.
Overall, multimodal fusion improves in-domain closed-set recognition but provides no consistent advantage in anomaly or open-set segmentation. These results expose a mismatch between known-class recognition, anomaly ranking, and thresholded open-set prediction.

\begin{table*}[htbp]
\centering
\scriptsize
\caption{\textbf{3D open-set semantic segmentation on DriveOSS-OOD test.}
We report class-wise open-IoU for the 19 ID classes, outlier IoU, and
mOpen-IoU over ID classes. M denotes the input modality. M4F3D* is the
modified LiDAR--camera fusion model.}
\label{tab:openset_classwise_test}
\resizebox{\textwidth}{!}{
\begin{tabular}{l c c c c c c c c c c c c c c c c c c c c c c}
\toprule
Method & M
& \rotatebox{90}{car}
& \rotatebox{90}{bicycle}
& \rotatebox{90}{motorcycle}
& \rotatebox{90}{truck}
& \rotatebox{90}{other-veh.}
& \rotatebox{90}{person}
& \rotatebox{90}{bicyclist}
& \rotatebox{90}{motorcyclist}
& \rotatebox{90}{road}
& \rotatebox{90}{parking}
& \rotatebox{90}{sidewalk}
& \rotatebox{90}{other-ground}
& \rotatebox{90}{building}
& \rotatebox{90}{fence}
& \rotatebox{90}{vegetation}
& \rotatebox{90}{trunk}
& \rotatebox{90}{terrain}
& \rotatebox{90}{pole}
& \rotatebox{90}{traffic-sign}
& \rotatebox{90}{outlier}
& \rotatebox{90}{mOpen-IoU} \\

\midrule
P3Former + MSP & L
& 24.69 & 0.0 & 0.0 & 6.56 & 0.0
& 0.14 & 1.00 & 0.0
& 0.03 & 0.0 & 0.0 & 0.0
& 0.05 & 0.13 & 1.26 & 0.05
& 0.0 & 0.03 & 0.06
&0.28  & 1.79 \\

P3Former + ML & L
& 22.96 & 0.0 & 0.0 & 6.33 &0.0
&0.95  & 0.28 & 0.0
& 0.02 & 0.0 & 0.0 & 0.0
& 1.37 & 1.08 & 0.48
& 0.28 & 0.01 & 0.07
& 0.27 & 0.27 & 1.80\\

P3Former + Energy & L
& 25.45 & 0.0 & 0.0 & 6.55 & 0.0
& 1.14 & 0.30 & 0.0
& 0.02 & 0.0 & 0.0 & 0.0
& 1.70 & 1.16 & 0.49 & 0.32
& 0.0 & 0.07 & 0.30
&0.27  &1.98  \\

P3Former + Entropy & L
& 24.65 & 0.0 & 0.0 & 6.63 & 0.0
& 0.15 & 0.85 & 0.0
& 0.02 & 0.0 & 0.0 & 0.0
& 0.05 & 0.16 & 1.07 & 0.05
& 0.0 & 0.03 & 0.05
& 0.27 & 1.78 \\

P3Former + RbA & L
& 2.13 & 0.0 & 0.0 & 0.0 & 0.0
& 0.0 & 0.0 & 0.0
& 16.62 & 1.03 & 0.51 & 0.72
& 65.18 & 53.47 & 0.08 & 13.10
& 4.49 & 0.38 & 1.14
& 0.51 & 8.36 \\

\midrule

LCPS-lidar + MSP & L
& 41.78 & 9.74 & 1.46 & 11.21 & 0.0
& 29.56 & 9.91 & 0.0
& 31.21 & 0.13 & 8.55 & 0.0
& 74.63 & 35.17 & 54.87 & 5.60
& 0.33 & 13.55 & 64.90
& 0.61 & 20.66 \\

LCPS-lidar + ML & L
& 16.48 & 17.52 & 2.54 & 11.19 & 0.0
& 23.98 & 16.67 & 0.0
& 16.40 & 0.93 & 35.03 & 0.29
& 40.60 & 19.63 & 16.93 & 19.51
& 1.68 & 13.03 & 31.90
&0.25  &14.97  \\

LCPS-lidar + Energy & L
& 16.33 & 17.29 &2.48  &10.97  &0.0
& 23.71 & 16.46 & 0.0
& 16.03 & 0.92 & 34.12 & 0.29
& 39.99 & 18.82 & 16.50 & 18.68
& 1.63 & 12.53 & 31.39
& 0.27 & 14.64 \\

LCPS-lidar + Entropy & L
& 41.47 & 9.79 & 1.53 & 11.27 & 0.0
& 29.53 & 10.01 & 0.0
& 31.51 & 0.14 & 8.84 & 0.0
& 75.30 & 34.83 & 55.06 & 5.89
& 0.32 & 13.56 & 65.25
&0.61  &20.75  \\

LCPS-lidar + RbA & L
& 8.79 & 3.16 & 1.96 & 0.70 & 0.0
& 7.15 & 10.06 & 0.0
& 4.66 & 0.79 & 0.50 & 0.59
& 1.95 & 0.91 & 0.35 & 1.85
& 0.98 & 0.47 & 32.62
&0.32  &4.08  \\

\midrule M4F3D + MSP & L & 38.96 & 7.50 & 11.90 & 14.49 & 0.00 & 35.85 & 18.18 & 0.00 & 0.27 & 0.08 & 0.26 & 0.01 & 0.18 & 0.23 & 0.08 & 0.34 & 0.01 & 0.03 & 0.25 & 0.30 & 6.77 \\

M4F3D + ML & L & 28.77 & 6.83 & 8.93 & 9.82 & 0.00 & 30.10 & 16.11 & 0.00 & 3.71 & 0.26 & 2.51 & 0.02 & 7.19 & 1.43 & 7.68 & 4.52 & 0.52 & 2.97 & 3.72 & 0.33 & 7.11 \\

M4F3D + Energy & L & 16.69 & 6.56 & 6.26 & 7.35 & 0.09 & 20.36 & 8.59 & 0.00 & 1.60 & 0.54 & 0.84 & 0.13 & 1.20 & 0.34 & 1.12 & 0.83 & 0.19 & 0.41 & 0.59 & 0.32 & 3.88 \\

M4F3D + Entropy & L & 34.41 & 8.15 & 11.91 & 14.26 & 0.05 & 34.42 & 18.94 & 0.00 & 0.25 & 0.08 & 0.25 & 0.01 & 0.16 & 0.26 & 0.05 & 0.29 & 0.01 & 0.03 & 0.27 & 0.30 & 6.52 \\

M4F3D + RbA & L & 24.90 & 11.11 & 8.39 & 11.84 & 0.53 & 28.14 & 27.06 & 0.00 & 18.85 & 13.46 & 17.87 & 2.43 & 4.44 & 2.64 & 0.95 & 3.59 & 3.43 & 0.73 & 1.60 & 0.32 & 9.58 \\

\midrule LIDO & L & 38.77 & 7.64 & 0.00 & 0.00 & 0.32 & 18.52 & 17.06 & 0.00 & 38.76 & 11.97 & 26.27 & 0.02 & 93.97 & 62.56 & 78.08 & 13.96 & 21.36 & 7.30 & 0.16 & 1.07 & 22.98 \\

\midrule
LCPS + MSP & L+C
& 37.99 & 1.91 & 3.56 & 7.22 & 0.22
& 24.57 & 17.38 & 0.0
& 25.86 & 0.09 & 5.41 & 0.04
& 76.74 & 27.13 & 40.03 & 4.51
& 1.65 & 8.41 & 45.23
& 0.57 & 17.26 \\

LCPS + ML & L+C
& 4.99 & 0.59 & 0.0 & 0.89 & 0.0
& 6.15 & 4.20 & 0.0
& 11.45 & 0.97 & 0.67 & 0.66
& 26.86 & 6.04 & 18.70 & 4.97
& 5.61 & 2.97 & 5.58
&0.39  & 5.33 \\

LCPS + Energy & L+C
& 4.78 & 0.49 & 0.0 & 0.07 & 0.0
& 5.67 & 3.74 & 0.0
& 11.20 & 0.95 & 0.62 & 0.65
& 25.97 & 5.80 & 18.21 & 4.78
& 5.52 & 2.93 & 5.31
& 0.39 & 5.09 \\

LCPS + Entropy & L+C
& 37.88 & 1.94 & 3.60 & 7.40 & 0.22
& 24.64 & 17.50 & 0.0
& 26.08 & 0.10 & 5.76 & 0.04
& 77.60 & 26.88 & 40.30 & 4.71
& 1.67 & 8.49 & 45.93
& 0.57 & 17.41 \\

LCPS + RbA & L+C
&  4.73& 4.47 & 5.03 & 11.37 & 0.04
& 19.68 & 22.66 & 0.0
& 11.50 & 2.18 & 10.25 & 2.05
& 20.38 & 1.98 & 14.04 & 32.98
& 9.24 & 2.28 & 7.48
&0.37  &9.60  \\

\midrule
M4F3D* + MSP & L+C & 53.49 & 2.51 & 6.67 & 8.42 & 0.53 & 32.45 & 28.43 & 0.56 & 10.84 & 0.00 & 0.56 & 0.05 & 1.19 & 0.66 & 0.78 & 2.00 & 0.46 & 0.33 & 0.52 & 0.30 & 7.92 \\

M4F3D* + ML & L+C & 26.19 & 1.27 & 3.45 & 6.96 & 0.10 & 17.83 & 14.10 & 0.00 & 2.17 & 0.00 & 0.30 & 0.01 & 0.37 & 0.11 & 0.19 & 0.50 & 0.03 & 0.04 & 0.07 & 0.31 & 3.88 \\

M4F3D* + Energy & L+C & 12.69 & 0.82 & 1.04 & 4.72 & 0.14 & 8.53 & 10.76 & 0.00 & 0.28 & 0.00 & 0.06 & 0.01 & 0.01 & 0.00 & 0.00 & 0.10 & 0.00 & 0.01 & 0.03 & 0.32 & 2.06 \\

M4F3D* + Entropy & L+C & 51.67 & 2.60 & 6.62 & 9.04 & 0.60 & 31.04 & 29.93 & 0.51 & 9.76 & 0.00 & 0.58 & 0.05 & 1.14 & 0.50 & 0.69 & 1.77 & 0.40 & 0.26 & 0.47 & 0.30 & 7.77 \\

M4F3D* + RbA & L+C & 44.83 & 5.30 & 4.88 & 8.21 & 2.48 & 37.81 & 32.96 & 0.52 & 17.18 & 0.03 & 18.55 & 0.65 & 10.49 & 2.91 & 3.45 & 8.86 & 7.87 & 2.94 & 4.20 & 0.21 & 11.27 \\

\bottomrule
\end{tabular}
}
\end{table*}

\begin{table*}[!hbtp]
\centering
\caption{\textbf{Inference time on the DriveOSS-OOD test set.} Per-frame runtime in ms, measured on a single NVIDIA RTX 4090 GPU and decomposed into the base segmentation network and post-hoc anomaly scoring overhead. Since LIDO jointly performs semantic and anomaly segmentation, only its end-to-end runtime is reported.}
\label{tab:inference_time}
\resizebox{0.80\linewidth}{!}{%
\begin{tabular}{lc|c|ccccc|c}
\toprule
& & \textbf{Segmentation} & \multicolumn{5}{c|}{\textbf{Post-hoc Scoring Overhead}} & \textbf{Total Time} \\
\textbf{Method} & \textbf{M} & \textbf{Network}  & +MSP & +ML & +Energy & +Entropy & +RbA & (ms) \\
\midrule
P3Former   & L   & 51.29  & 0.07 & 0.10 & 0.13 & 0.10 & 0.11 & [51.36, 51.42] \\
LCPS-lidar & L   & 85.44  & 0.12 & 0.17 & 0.25 & 0.19 & 0.18 & [85.56, 85.69] \\
M4F3D      & L   & 104.21 & 0.11 & 0.17 & 0.23 & 0.18 & 0.13 & [104.32, 104.44] \\
LIDO       & L   & --    & --  & --  & --  & --  & --  & 68.01 \\
\midrule
LCPS       & L+C & 179.80 & 0.11 & 0.16 & 0.21 & 0.15 & 0.16 & [179.91, 180.01] \\
M4F3D*     & L+C & 106.35 & 0.10 & 0.16 & 0.21 & 0.17 & 0.12 & [106.45, 106.56] \\
\bottomrule
\end{tabular}%
}
\end{table*}

\paragraph{Inference Efficiency.}
As shown in Table~\ref{tab:inference_time}, post-hoc anomaly scoring functions add at most 0.25\,ms per frame, indicating that runtime is dominated by the segmentation network. P3Former is the fastest baseline at approximately 51.4\,ms per frame, while LIDO jointly performs semantic and anomaly segmentation in 68.0\,ms. Among the multimodal models, M4F3D* requires 106.5\,ms per frame, substantially less than the 180.0\,ms required by LCPS. Specifically, M4F3D* adds only 2.14\,ms (2.05\%) over M4F3D, whereas LCPS adds 94.36\,ms (110.44\%) over LCPS-lidar. This comparison suggests that query-level cross-modal interaction can be substantially more efficient than dense point--image fusion. However, the low fusion overhead of M4F3D* does not consistently translate into improved anomaly or open-set performance, showing that computational efficiency alone does not ensure robust ID--OOD separation.

\section{Conclusion}
\label{conclusion}

In this work, we introduce DriveOSS, which, to the best of our knowledge, is the first real-world dataset specifically designed for 3D open-set semantic segmentation. It provides synchronized multimodal data collected across diverse locations, weather conditions, and illumination settings, with point-wise annotations for both known classes and OOD objects. We adapt the open-IoU protocol to 3D point clouds and benchmark LiDAR-only and LiDAR-camera fusion models using post-hoc anomaly scoring methods, together with the feature-based method. The results show that current approaches still struggle to detect unknown objects while preserving accurate semantic predictions for known classes, highlighting the challenges of real-world 3D open-set perception.

\section{Limitations and Future Work}
\label{limitation}

The current benchmark focuses on single-frame perception using a fixed sensor configuration and does not explicitly exploit temporal information. Future work will investigate temporal and unified open-set segmentation models that jointly optimize known-class recognition and unknown-object detection. We also plan to extend DriveOSS to open-set occupancy prediction, enabling voxel-level reasoning over free space, known classes, and unknown objects.

\section{Acknowledgment}
\label{ACKNOWLEDGMENT}
The authors would like to thank the Federal Ministry for
Economic Affairs and Climate Action for supporting this work
under the project “Gaia-X 4 AMS”.

{
    \small
    \bibliographystyle{ieeenat_fullname}
    \bibliography{references}
}

\clearpage
\maketitlesupplementary
\appendix
\setcounter{figure}{0}
\setcounter{table}{0}
\setcounter{equation}{0}
\renewcommand{\thefigure}{S\arabic{figure}}
\renewcommand{\thetable}{S\arabic{table}}
\renewcommand{\theequation}{S\arabic{equation}}
\renewcommand{\theHfigure}{supp.\arabic{figure}}
\renewcommand{\theHtable}{supp.\arabic{table}}
\renewcommand{\theHequation}{supp.\arabic{equation}}

\section{Data Acquisition}
\label{data_acquisition_suppl}

\subsection{Sensor Setup}
DriveOSS was collected using a vehicle equipped with one LiDAR, six surround-view RGB cameras, and one GPS/IMU system. Table~\ref{tab:sensor_specification} summarizes the sensor specifications.

\begin{table*}[t]
\centering
\small
\caption{\textbf{Sensor specifications of the DriveOSS platform.}}
\label{tab:sensor_specification}
\begin{tabularx}{\textwidth}{l l c X}
\toprule
\textbf{Sensor Type} & \textbf{Sensor Model} & \textbf{Number} & \textbf{Details} \\
\midrule
LiDAR & Ouster OS1-128 & $\times$1 & 128 beams, $360^\circ$ horizontal FOV, $-22.5^\circ$ to $+22.5^\circ$ vertical FOV, 20 Hz \\
Camera & Basler ace2R a2A1920-51gcBAS & $\times$6 & RGB, $1920 \times 1080$ resolution, 10 Hz \\
GPS/IMU & Certus & $\times$1 & $>$600 Hz update rate, dual-antenna GNSS/INS, 0.01 m horizontal accuracy with RTK, 0.015 m vertical accuracy with RTK, and 0.1$^\circ$ roll/pitch/heading accuracy \\
\bottomrule
\end{tabularx}
\end{table*}

\subsection{Privacy Protection}
Before public release, faces and license plates are anonymized using a two-stage procedure. A detection model first identifies and blurs sensitive regions, followed by manual frame-level verification. This process reduces privacy risks and supports responsible public release.

\section{Evaluation Protocols}
\label{tasks_suppl}

\subsection{Closed-Set 3D Semantic Segmentation}
\textbf{Evaluation Metrics.}
Closed-set semantic segmentation evaluates prediction quality only on ID classes. Points labeled as outlier ($y_i = K+1$) or ignored ($y_i = 0$) are excluded from evaluation. For each class $k \in \mathcal{Y}_{\mathrm{ID}}$, the intersection-over-union (IoU) is defined as
\begin{equation}
\mathrm{IoU}_k = \frac{\mathrm{TP}_k}{\mathrm{TP}_k + \mathrm{FP}_k + \mathrm{FN}_k}.
\end{equation}
The mean IoU (mIoU)~\cite{blum2021fishyscapes} is computed as
\begin{equation}
\mathrm{mIoU} = \frac{1}{K} \sum_{k=1}^{K} \mathrm{IoU}_k.
\end{equation}

\subsection{3D Anomaly Segmentation}

\textbf{Evaluation Metrics.}
Anomaly segmentation is formulated as a binary discrimination task between ID and OOD points. The binary ground-truth label is derived from the semantic label $y_i$ as
\begin{equation}
z_i =
\begin{cases}
1, & y_i = K+1 \quad (\text{OOD}),\\
0, & y_i \in \{1, \dots, K\} \quad (\text{ID}),\\
-1, & y_i = 0 \quad (\text{ignored}).
\end{cases}
\end{equation}
Points with $z_i = -1$ are excluded from evaluation.

The model output $s_i$ is a continuous score rather than a discrete prediction. Therefore, performance is evaluated based on how well the scores $s_i$ separate OOD and ID points. We adopt common metrics~\cite{blum2021fishyscapes} for point-level evaluation. Specifically, AUROC and AP measure the ranking quality of anomaly scores across all possible thresholds, while FPR95 reports the false positive rate when the true positive rate reaches 95\%.

\subsection{3D Open-Set Semantic Segmentation}

\textbf{Task Definition.}
Open-set segmentation extends semantic segmentation by jointly recognizing ID classes and detecting OOD points. Given a LiDAR point cloud $\mathcal{X} = \{p_i\}_{i=1}^N$ (optionally with corresponding RGB images), the model predicts a label $\hat{y}_i \in \{1, \dots, K+1\}$ for each point, where $\{1, \dots, K\}$ denote ID classes and the $(K+1)$-th label corresponds to the outlier class.

\textbf{Evaluation Metrics.}
Let $\hat{y}_i^{\mathrm{cls}}$ denote the closed-set semantic prediction and $s_i$ the anomaly score, where larger values indicate a higher likelihood of OOD. For each method, $\tau_{95}$ is calibrated on DriveOSS-OOD validation such that the OOD true positive rate reaches 95\%, and is then fixed for test-set evaluation. The final prediction is
\begin{equation}
\hat{y}_i^{\mathrm{open}} =
\begin{cases}
K+1, & s_i > \tau_{95},\\
\hat{y}_i^{\mathrm{cls}}, & s_i \leq \tau_{95}.
\end{cases}
\end{equation}

We adopt open-set intersection-over-union (open-IoU)~\cite{grcic2022densehybrid}. For each ID class $k \in \mathcal{Y}_{\mathrm{ID}}$, it is computed from the final open-set predictions as
\begin{equation}
\mathrm{open\text{-}IoU}_k =
\frac{\mathrm{TP}_k}
{\mathrm{TP}_k + \mathrm{FP}_k^{\mathrm{ow}} + \mathrm{FN}_k^{\mathrm{ow}}},
\end{equation}
where $\mathrm{FP}_k^{\mathrm{ow}}$ and $\mathrm{FN}_k^{\mathrm{ow}}$ are computed over all $K+1$ output classes, including \emph{outlier}. Rejecting an ID point as OOD contributes a false negative to its semantic class, while assigning an OOD point to class $k$ contributes a false positive. We additionally report outlier IoU separately.

The final performance is reported as the mean open-IoU over ID classes:
\begin{equation}
\mathrm{mOpen\text{-}IoU} = \frac{1}{K} \sum_{k=1}^{K} \mathrm{open\text{-}IoU}_k.
\end{equation}

\section{Query-Level Camera-LiDAR Fusion}
\label{sec:query_fusion}
We extend Mask4Former3D with a camera branch and perform fusion at the decoder-query level. Given the synchronized front-view image, a ResNet-18 backbone up to \texttt{layer3} extracts 2D features, followed by a $1\times1$ projection to the decoder dimension. The feature map is flattened into image tokens and augmented with fixed 2D sine--cosine positional encodings.

The sparse 3D backbone and query initialization remain unchanged. Decoder queries are initialized from farthest-point-sampled 3D coordinates. At each decoder stage, we insert camera cross-attention after LiDAR cross-attention, followed by self-attention and feed-forward refinement.

Let $\mathbf{Q}^{(l)}$ be the queries at stage $l$, $\mathbf{T}^{\text{pcd},(l)}$ the LiDAR memory, and $\mathbf{T}^{\text{img}}$ the image tokens. The update is defined as
\[
\hat{\mathbf{Q}}^{(l)} = \mathrm{CA}_{\text{pcd}}(\mathbf{Q}^{(l)}, \mathbf{T}^{\text{pcd},(l)}), \quad
\bar{\mathbf{Q}}^{(l)} = \mathrm{CA}_{\text{img}}(\hat{\mathbf{Q}}^{(l)}, \mathbf{T}^{\text{img}}),
\]
\[
\mathbf{Q}^{(l+1)} = \mathrm{FFN}\big(\mathrm{SA}(\bar{\mathbf{Q}}^{(l)})\big).
\]
Residual connections and normalization are omitted above for clarity. The design preserves the LiDAR-centric representation while injecting complementary appearance information. Because interaction occurs only through decoder queries, it requires no explicit point-to-pixel projection and adds little latency over the LiDAR-only model.

\section{Implementation Details}

All models are trained on ID data only. The joint training set comprises SemanticKITTI sequences \{00--07, 09, 10\}, Panoptic-CUDAL sequences \{30, 31, 36, 40, 41\}, and DriveOSS-ID sequences \{311--315, 321, 322, 324, 327\}. DriveOSS-ID sequences \{316--320, 323, 325, 326, 329, 330\} form the closed-set validation set. Every weather condition is represented in both DriveOSS-ID splits.

DriveOSS-OOD validation contains sequences \{411, 412, 421, 422, 431, 432, 441, 442, 451, 452, 461--467\}; test contains \{413, 414, 423, 424, 433, 434, 443, 444, 453, 454, 471--476\}. The two OOD splits were recorded at geographically disjoint locations. OOD samples and anomaly annotations are never used for training.

\textbf{P3Former.} We retain the original architecture and hyperparameters. The model is trained for 36 epochs on two RTX 4090 GPUs with batch size 1 and learning rate $0.008$.

\textbf{LCPS.} We evaluate both LCPS-lidar and the LiDAR--camera LCPS model. Because the unified training setup provides one front-view image per frame, the fusion model is adapted from the original two-view configuration to a single front view. Instance augmentation is disabled because the required instance-level annotations are unavailable for all training datasets. Both variants are trained for 30 epochs with batch size 1 and learning rate $0.004$.

\textbf{Mask4Former3D (M4F3D).} We follow the original architecture and training configuration. The model is trained for 30 epochs on two RTX 4090 GPUs with batch size 2 and learning rate $0.008$.

\textbf{Mask4Former3D* (M4F3D*).} M4F3D* augments M4F3D with the query-level camera branch described in Sec.~\ref{sec:query_fusion} and takes the front-view image and LiDAR point cloud as input. It uses the same 30-epoch schedule, batch size 2, and learning rate $0.008$ on two RTX 4090 GPUs.

\textbf{LIDO.} We use the official feature-based semantic and anomaly segmentation framework~\cite{Mosco_2026_CVPR} and train it on the same unified ID-only data. No DriveOSS-OOD frames or anomaly annotations are used. Its semantic predictions are evaluated for closed-set segmentation, while its feature-based anomaly scores are used for anomaly and open-set evaluation. It is trained on the unified ID-only data
for 64 epochs on two RTX 4090 GPUs, using batch size 8 and learning
rate $2.4\times10^{-1}$.

\section{Additional Results}

We provide complete class-wise closed-set results on DriveOSS-ID validation, DriveOSS-OOD validation, and DriveOSS-OOD test in Tables~\ref{tab:closed_semantic_classwise_ID_val}--\ref{tab:closed_semantic_classwise_OOD_test}. Tables~\ref{tab:anomaly_val} and~\ref{tab:openset_classwise_val} report anomaly and open-set semantic segmentation on DriveOSS-OOD validation.

\begin{table*}[hbtp]
\centering
\scriptsize
\caption{Closed-set 3D semantic segmentation results on the DriveOSS-ID validation set. Class-wise IoU and mIoU are reported. ``M'' denotes the input modality, where ``L'' and ``C'' represent LiDAR and camera, respectively. M4F3D* denotes the modified LiDAR--camera fusion model.}
\label{tab:closed_semantic_classwise_ID_val}
\resizebox{\textwidth}{!}{
\begin{tabular}{l c c c c c c c c c c c c c c c c c c c c c}
\toprule
Method & M
& \rotatebox{90}{car}
& \rotatebox{90}{bicycle}
& \rotatebox{90}{motorcycle}
& \rotatebox{90}{truck}
& \rotatebox{90}{other-veh.}
& \rotatebox{90}{person}
& \rotatebox{90}{bicyclist}
& \rotatebox{90}{motorcyclist}
& \rotatebox{90}{road}
& \rotatebox{90}{parking}
& \rotatebox{90}{sidewalk}
& \rotatebox{90}{other-ground}
& \rotatebox{90}{building}
& \rotatebox{90}{fence}
& \rotatebox{90}{vegetation}
& \rotatebox{90}{trunk}
& \rotatebox{90}{terrain}
& \rotatebox{90}{pole}
& \rotatebox{90}{traffic-sign}
& \rotatebox{90}{mIoU} \\
\midrule

P3Former & L
& 83.03 & 1.45 & 0.20& 45.10&44.08 &24.12 &30.29 &4.65 &64.77
&30.19 &19.69 &2.34 & 91.31&57.76 &87.92 &65.99 &46.78
&57.73 &64.64 & 43.27\\

LCPS-lidar & L
&85.52 &23.06 &8.15 &48.59 &41.16 &34.51 &54.58 &21.42 &79.49
&34.22 &59.63 &2.62 &94.51 &47.55 &91.55 &70.93 &54.05
&58.69 &76.60 &51.94 \\

M4F3D & L
& 73.96 & 6.51 & 22.04 & 8.69 & 26.69
& 15.52 & 40.65 & 1.86
& 63.21 & 22.79 & 17.01 & 2.29
& 66.02 & 42.00 & 87.04 & 43.74
& 40.38 & 34.53 & 49.69 & 34.98 \\

LIDO & L & 90.93 & 19.87 & 29.64 & 73.91 & 63.43 & 51.10 & 80.35 & 25.32 & 81.29 & 37.98 & 57.06 & 8.63 & 93.52 & 55.53 & 88.91 & 70.43 & 56.49 & 50.39 & 73.92 & 58.35 \\

\midrule

LCPS & L+C
& 91.80& 15.22& 27.37& 50.12& 64.59& 31.53&55.11 &12.66 &78.32
&32.91 &56.54 &2.58 &96.29 &55.64 &91.73 &71.48 &57.63
&64.19 &71.18 &54.05 \\

M4F3D* & L+C
& 84.25 & 9.41 & 27.01 & 31.98 & 20.73
& 46.02 & 68.33 & 42.07
& 72.51 & 16.85 & 48.63 & 1.69
& 86.78 & 44.35 & 88.61 & 67.21
& 43.50 & 49.85 & 70.55 & 48.44 \\

\bottomrule
\end{tabular}
}
\end{table*}

\begin{table*}[htbp]
\centering
\scriptsize
\caption{Closed-set 3D semantic segmentation results on the DriveOSS-OOD validation set.
Class-wise IoU and mIoU over the 19 known classes are reported.
``M'' denotes the input modality, where ``L'' and ``C'' indicate LiDAR and camera, respectively. M4F3D* represents the modified fusion model.}
\label{tab:semantic_classwise_val}
\resizebox{\textwidth}{!}{
\begin{tabular}{l c c c c c c c c c c c c c c c c c c c c c}
\toprule
Method & M
& \rotatebox{90}{car}
& \rotatebox{90}{bicycle}
& \rotatebox{90}{motorcycle}
& \rotatebox{90}{truck}
& \rotatebox{90}{other-veh.}
& \rotatebox{90}{person}
& \rotatebox{90}{bicyclist}
& \rotatebox{90}{motorcyclist}
& \rotatebox{90}{road}
& \rotatebox{90}{parking}
& \rotatebox{90}{sidewalk}
& \rotatebox{90}{other-ground}
& \rotatebox{90}{building}
& \rotatebox{90}{fence}
& \rotatebox{90}{vegetation}
& \rotatebox{90}{trunk}
& \rotatebox{90}{terrain}
& \rotatebox{90}{pole}
& \rotatebox{90}{traffic-sign}
& \rotatebox{90}{mIoU} \\
\midrule
P3Former & L
&81.50  & 0.20 & 0.13 &18.31 &37.36 &33.70
&2.81  &2.76  &52.11
& 4.80 & 6.48 & 0.34 & 91.88
& 59.06 & 85.07 &75.38  &28.51
&36.49 & 46.12& 34.90   \\

M4F3D & L
& 85.28 & 30.02 & 3.47  & 59.85 & 54.47
& 75.57 & 31.46 & 23.47
& 49.90 & 12.59 & 14.42 & 6.98
& 91.26 & 40.31 & 85.04 & 75.99
& 33.98 & 32.90 & 62.08 & 45.74 \\

LCPS-lidar & L
& 86.02 & 5.02 & 0.0 & 64.99& 13.03& 53.16
& 21.61 & 5.74 & 61.36
& 9.18 & 10.41 & 4.42 & 91.47
& 48.72 & 86.15 & 76.54 & 34.45
& 42.20 & 57.10 & 40.61    \\

LIDO & L & 88.95 & 23.22 & 3.41 & 73.95 & 20.92 & 77.69 & 40.93 & 19.48 & 55.17 & 29.29 & 15.71 & 12.01 & 92.14 & 36.93 & 84.56 & 65.24 & 40.66 & 27.23 & 58.64 & 45.59 \\

\midrule
LCPS & L+C
& 86.64 & 1.48 & 0.48 & 36.23& 21.75& 50.84
& 12.73 & 4.12 &58.26
& 14.20 & 8.68 & 6.39 & 92.74
& 51.89 & 86.42 & 78.06 & 33.84
&35.93 & 55.99& 38.78    \\
M4F3D* & L+C & 73.57 & 3.28 & 10.76 & 10.42 & 24.50 & 59.25 & 0.00 & 18.53 & 45.59 & 0.07 & 7.99 & 2.11 & 77.74 & 25.48 & 85.07 & 67.30 & 29.49 & 23.71 & 34.76 & 31.56 \\

\bottomrule
\end{tabular}
}
\end{table*}

\begin{table*}[hbtp]
\centering
\scriptsize
\caption{Closed-set 3D semantic segmentation results on the DriveOSS-OOD test set. Class-wise IoU and mIoU are reported. "M" means modality, and "L" and "C" denote LiDAR and Camera, respectively. M4F3D* represents the modified fusion model.}
\label{tab:closed_semantic_classwise_OOD_test}
\resizebox{\textwidth}{!}{
\begin{tabular}{l c c c c c c c c c c c c c c c c c c c c c}
\toprule
Method & M
& \rotatebox{90}{car}
& \rotatebox{90}{bicycle}
& \rotatebox{90}{motorcycle}
& \rotatebox{90}{truck}
& \rotatebox{90}{other-veh.}
& \rotatebox{90}{person}
& \rotatebox{90}{bicyclist}
& \rotatebox{90}{motorcyclist}
& \rotatebox{90}{road}
& \rotatebox{90}{parking}
& \rotatebox{90}{sidewalk}
& \rotatebox{90}{other-ground}
& \rotatebox{90}{building}
& \rotatebox{90}{fence}
& \rotatebox{90}{vegetation}
& \rotatebox{90}{trunk}
& \rotatebox{90}{terrain}
& \rotatebox{90}{pole}
& \rotatebox{90}{traffic-sign}
& \rotatebox{90}{mIoU} \\
\midrule

P3Former & L
& 61.64 & 0.34 & 0.15 & 7.42 & 3.22
& 22.47 & 14.27 & 0.0
& 53.58 & 1.88 & 13.27 & 3.03
& 94.60 & 53.74 & 83.44 & 46.87
& 29.07 & 18.84 & 54.13 & 29.58  \\

LCPS-lidar & L
& 61.46 & 21.19 & 2.83 & 13.49 & 0.0
& 36.86 & 17.84 & 0.0
& 54.59 & 6.14 & 19.48 & 2.71
& 95.63 & 52.44 & 85.76 & 55.92
& 22.49 & 20.31 & 72.12 & 33.75 \\

M4F3D & L & 59.62 & 11.83 & 8.66 & 10.02 & 0.60 & 54.11 & 43.22 & 0.00 & 44.65 & 16.62 & 26.85 & 3.19 & 82.82 & 33.25 & 80.63 & 38.60 & 13.87 & 9.41 & 53.19 & 31.11 \\

LIDO & L & 68.18 & 28.94 & 0.55 & 15.03 & 36.02 & 62.09 & 58.62 & 3.47 & 56.52 & 25.19 & 22.48 & 14.61 & 96.77 & 45.39 & 85.76 & 42.54 & 34.06 & 12.29 & 70.31 & 40.99 \\

\midrule
LCPS & L+C
& 60.91 & 9.81 & 7.50 & 14.03 & 0.43
& 33.93 & 24.87 & 0.0
& 52.71 & 14.87 & 22.10 & 6.32
& 96.74 & 58.67 & 85.36 & 53.60
& 28.23 & 16.54 & 59.30 & 34.00 \\

M4F3D* & L+C & 60.58 & 5.40 & 7.80 & 7.66 & 2.52 & 51.12 & 35.81 & 0.50 & 43.87 & 0.03 & 18.70 & 0.70 & 83.11 & 25.50 & 80.53 & 38.48 & 14.26 & 12.64 & 51.56 & 28.46 \\
\bottomrule
\end{tabular}
}
\end{table*}

\begin{table}[htbp]
\centering
\small
\caption{3D anomaly segmentation results on the DriveOSS-OOD validation set. We report AUROC~$\uparrow$, FPR@95~$\downarrow$, and AP~$\uparrow$. M4F3D* represents the modified fusion model.}
\label{tab:anomaly_val}
\resizebox{\columnwidth}{!}{
\begin{tabular}{l c c c c}
\toprule
Method & M & AUROC~$\uparrow$ & FPR@95~$\downarrow$ & AP~$\uparrow$ \\
\midrule

P3Former + MSP
& L & 35.34 & 99.02 & 0.13 \\

P3Former + ML
& L & 41.72 & 98.55 & 0.14 \\

P3Former + Energy
& L & 46.53 & 98.42 & 0.26 \\

P3Former + Entropy
& L & 32.64 & 99.03 & 0.14 \\

P3Former + RbA
& L & 86.43 & 61.27 & 1.57 \\

\midrule
LCPS-lidar + MSP
& L & 72.51 & 55.61 & 0.31 \\

LCPS-lidar + ML
& L & 65.90 & 84.55 & 0.28 \\

LCPS-lidar + Energy
& L & 65.57 &84.84 & 0.27 \\

LCPS-lidar + Entropy
& L & 73.80 & 55.33 & 0.42 \\

LCPS-lidar + RbA
& L & 48.37 &96.36 & 0.12 \\

\midrule
M4F3D + MSP
& L & 83.69 & 98.40 & 8.16 \\

M4F3D + ML
& L & 78.99 & 87.29 & 7.98 \\

M4F3D + Energy
& L & 72.27 & 94.67 & 7.96 \\

M4F3D + Entropy
& L & 81.00 & 98.62 & 8.84 \\

M4F3D + RbA
& L & 66.90 & 100.00 & 9.75 \\
\midrule

LIDO
& L & 92.97 & 30.13 & 3.35\\

\midrule
LCPS + MSP
& L+C & 74.29 & 56.89 & 0.40 \\

LCPS + ML
& L+C & 67.51 & 82.27 & 0.40 \\

LCPS + Energy
& L+C & 66.85 & 82.72 & 0.37 \\

LCPS + Entropy
& L+C & 75.69 & 56.49 & 0.70 \\

LCPS + RbA
& L+C & 71.96 & 80.55 & 1.33 \\

\midrule
M4F3D* + MSP
& L+C & 73.55 & 93.52 & 3.00 \\

M4F3D* + ML
& L+C & 69.66 & 93.75 & 2.97 \\

M4F3D* + Energy
& L+C & 60.11 & 95.20 & 2.67 \\

M4F3D* + Entropy
& L+C & 71.48 & 93.54 & 3.06 \\

M4F3D* + RbA
& L+C & 47.30 & 100.00 & 2.78 \\

\bottomrule
\end{tabular}
}
\vspace{-5mm}
\end{table}

\begin{table*}[htbp]
\centering
\scriptsize
\caption{3D open-set segmentation results on the DriveOSS-OOD validation set.
Class-wise open-IoU is reported for the 19 known classes, along with outlier IoU and mOpen-IoU over known classes.
"M" denotes the input modality: L for LiDAR-only and L+C for LiDAR-camera fusion. M4F3D* represents the modified fusion model.}
\label{tab:openset_classwise_val}
\resizebox{\textwidth}{!}{
\begin{tabular}{l c c c c c c c c c c c c c c c c c c c c c c}
\toprule
Method & M
& \rotatebox{90}{car}
& \rotatebox{90}{bicycle}
& \rotatebox{90}{motorcycle}
& \rotatebox{90}{truck}
& \rotatebox{90}{other-veh.}
& \rotatebox{90}{person}
& \rotatebox{90}{bicyclist}
& \rotatebox{90}{motorcyclist}
& \rotatebox{90}{road}
& \rotatebox{90}{parking}
& \rotatebox{90}{sidewalk}
& \rotatebox{90}{other-ground}
& \rotatebox{90}{building}
& \rotatebox{90}{fence}
& \rotatebox{90}{vegetation}
& \rotatebox{90}{trunk}
& \rotatebox{90}{terrain}
& \rotatebox{90}{pole}
& \rotatebox{90}{traffic-sign}
& \rotatebox{90}{outlier}
& \rotatebox{90}{mOpen-IoU} \\

\midrule
P3former + MSP & L& 20.46 & 0.0 & 0.08 & 17.26 & 0.73& 0.03 & 2.07 & 0.09 &0.05 & 0.0 & 0.0 & 0.0 & 0.01 & 0.03 & 0.15 & 0.0 & 0.0 & 0.0 & 0.82 & 0.10 & 2.20 \\
P3former + ML & L&  30.49& 0.0 & 0.16 & 16.68 &12.48 & 1.40 & 2.77 & 0.09 & 0.0& 0.0 & 0.0 & 0.0 & 0.44 & 3.58 & 0.10 & 0.0 & 0.01 & 0.01 & 0.53 & 0.10 & 3.62 \\
P3former + Energy & L & 33.03 & 0.0 & 0.15 & 17.08 & 14.80& 1.64 & 2.51 & 0.10 & 0.0& 0.0 &0.0  &0.0  &0.60  &4.35  &0.10  &0.0  &0.01  &0.01  &0.52  &0.10  & 3.94 \\
P3former + Entropy & L& 21.07 & 0.0 & 0.10 & 17.04 & 0.89& 0.05 & 3.52 & 0.10 & 0.02& 0.0 & 0.0 & 0.0 & 0.01 & 0.04 & 0.13 & 0.0 & 0.01 &0.0  &0.76  &0.10  &2.30  \\
P3former + RbA & L& 3.35 & 0.0 & 0.0 & 0.0 & 0.0& 0.0 & 0.0 & 0.0 & 34.14& 3.51 & 0.01 & 0.01 & 69.96 & 31.04 & 0.03 & 39.03 & 3.92 & 3.16 & 0.84 & 0.14 & 9.95 \\

\midrule
LCPS-lidar + MSP & L& 69.58 & 1.87 & 0.0 & 44.93 & 12.12& 42.17 & 25.34 & 5.18 & 29.48& 0.16 & 12.79 & 0.01 & 72.53 & 12.50 & 63.89 &21.76  &1.94  &35.77  &46.73  &0.17  & 26.25 \\
LCPS-lidar + ML & L& 17.40 & 0.23 & 0.0 & 24.42 & 0.05& 11.69 & 34.73 &0.0  &14.67 & 1.34 & 7.97 & 0.19 & 29.30 & 5.05 & 12.96 & 14.70 & 4.88 & 14.55 & 22.24 & 0.11 & 11.39 \\
LCPS-lidar + Energy & L & 17.07 & 0.20 & 0.0 & 23.80 & 0.05& 11.44 & 34.14 & 0.0 & 14.33& 1.31 & 7.54 & 0.20 & 28.70 & 4.65 & 12.64 & 14.18 & 4.65 & 13.83 & 21.70 & 0.11 & 11.07 \\
LCPS-lidar + Entropy & L& 69.36 & 1.86 & 0.0 & 44.96 &12.07 & 42.02 & 25.28 & 5.17 & 29.83& 0.16 & 13.00 & 0.01 & 73.04 & 12.83 & 64.05 & 21.94 & 1.93 & 35.08 & 46.73 & 0.17 & 26.28 \\
LCPS-lidar + RbA & L & 11.81 & 0.90 & 0.0 & 15.24 & 1.34& 14.70 & 12.31 & 4.39 &3.79 & 1.36 & 0.74 & 1.16 & 2.14 &0.65  &0.34  &0.80  &1.22  &0.40  &31.02  &0.09  & 5.49 \\

\midrule
M4F3D + MSP & L & 33.43 & 22.66 & 0.77 & 34.43 & 37.15 & 43.45 & 47.50 & 11.99 & 0.04 & 0.05 & 0.14 & 0.05 & 0.12 & 0.22 & 0.03 & 1.07 & 0.02 & 0.00 & 0.27 & 0.09 & 12.28 \\

M4F3D + ML & L & 31.78 & 17.93 & 2.47 & 40.65 & 36.20 & 37.90 & 43.21 & 12.64 & 10.47 & 0.94 & 4.64 & 0.27 & 11.17 & 18.66 & 18.59 & 12.21 & 2.96 & 11.11 & 9.97 & 0.11 & 17.04 \\

M4F3D + Energy & L & 16.29 & 14.88 & 4.06 & 31.41 & 28.99 & 26.41 & 33.58 & 15.59 & 4.61 & 1.44 & 1.87 & 0.85 & 2.34 & 4.40 & 5.98 & 4.03 & 1.77 & 3.06 & 3.00 & 0.10 & 10.77 \\

M4F3D + Entropy & L & 26.42 & 22.88 & 1.26 & 32.28 & 34.57 & 38.49 & 40.01 & 14.16 & 0.02 & 0.05 & 0.09 & 0.05 & 0.09 & 0.20 & 0.02 & 0.97 & 0.02 & 0.00 & 0.30 & 0.09 & 11.15 \\

M4F3D + RbA & L & 12.59 & 28.79 & 3.33 & 50.76 & 32.92 & 30.22 & 27.85 & 24.53 & 15.37 & 11.41 & 6.22 & 6.52 & 0.78 & 1.57 & 0.18 & 1.35 & 9.79 & 0.60 & 2.19 & 0.09 & 14.05 \\

\midrule LIDO & L & 67.57 & 2.07 & 0.00 & 0.03 & 0.13 & 33.34 & 15.99 & 6.64 & 42.17 & 18.38 & 18.68 & 0.02 & 89.16 & 46.33 & 75.90 & 31.23 & 25.46 & 14.92 & 0.02 & 0.31 & 25.69 \\

\midrule
LCPS + MSP & L+C
&  69.10& 0.14 &0.48  &11.78  &13.90 & 40.41 & 21.16 & 0.77 & 26.58& 0.10 &2.59  & 0.09 & 76.07 & 17.21 & 56.83 & 17.56 & 0.81 & 25.20 & 42.09 & 0.16 &22.26  \\

LCPS + ML & L+C
& 8.89 & 0.0 & 0.0 & 0.33 &1.93 &11.91  &0.91  & 3.73 & 10.34&  0.73&  0.52&  0.88&  31.14& 1.45 & 13.69 & 8.42 & 2.73 & 5.45 & 6.26 & 0.11 & 5.75 \\

LCPS + Energy & L+C
& 8.45 & 0.0 & 0.0 & 0.15 & 1.87& 11.29 & 0.60 & 3.65 & 10.12&  0.72& 0.50 & 0.87 & 30.28 & 1.38 & 13.31 & 8.03 & 2.67 & 5.22 & 6.00 & 0.10 &5.53  \\

LCPS + Entropy & L+C
& 69.08 & 0.15 & 0.48 & 12.03 &13.89 & 40.45 & 21.03 & 0.77 & 26.95& 0.11 & 2.75 & 0.10 & 76.75 & 17.66 & 57.21 & 18.06 & 0.85 & 24.82 & 42.62 & 0.16 & 22.41 \\

LCPS + RbA & L+C
& 2.69 & 0.31 & 0.04 & 25.64 & 0.63& 22.01 & 12.36 & 3.43 & 10.19& 2.21 & 4.93 & 1.44 & 18.02 & 1.66 & 14.31 & 38.44 & 9.51 & 3.18 & 9.73 & 0.11 & 9.51 \\

\midrule
M4F3D* + MSP & L+C & 58.21 & 0.56 & 4.84 & 29.17 & 22.63 & 36.35 & 0.00 & 24.97 & 9.75 & 0.00 & 0.21 & 0.01 & 1.00 & 0.23 & 0.55 & 2.02 & 0.43 & 0.26 & 0.41 & 0.10 & 10.08 \\

M4F3D* + ML & L+C & 57.22 & 0.62 & 3.40 & 27.97 & 20.02 & 34.13 & 0.00 & 24.51 & 9.10 & 0.00 & 0.49 & 0.02 & 1.43 & 3.00 & 1.18 & 2.24 & 0.49 & 0.84 & 1.06 & 0.10 & 9.88 \\

M4F3D* + Energy & L+C & 48.46 & 0.85 & 2.35 & 27.56 & 16.88 & 27.62 & 0.00 & 22.82 & 5.59 & 0.00 & 0.64 & 0.09 & 0.76 & 1.46 & 0.35 & 1.26 & 0.40 & 0.35 & 0.64 & 0.10 & 8.32 \\

M4F3D* + Entropy & L+C & 57.83 & 0.80 & 4.44 & 30.05 & 22.30 & 36.17 & 0.00 & 24.47 & 9.49 & 0.00 & 0.21 & 0.01 & 0.96 & 0.22 & 0.53 & 2.00 & 0.41 & 0.26 & 0.39 & 0.10 & 10.03 \\

M4F3D* + RbA & L+C & 52.47 & 3.24 & 9.87 & 13.76 & 21.77 & 48.36 & 0.00 & 18.85 & 13.01 & 0.07 & 6.48 & 2.04 & 3.31 & 5.77 & 1.52 & 6.31 & 16.91 & 1.93 & 5.11 & 0.07 & 12.15 \\

\bottomrule
\end{tabular}
}
\end{table*}

\section{Anomaly Scoring Methods}
\label{sec:anomaly_scoring}
This section introduces the anomaly scoring functions used in this work. Each method quantifies model uncertainty from a different perspective, allowing complementary cues for detecting out-of-distribution (OOD) regions.

\subsection{Maximum Softmax Probability (MSP)}
\textbf{Mathematical Formulation:}
\begin{equation}
U_{\text{MSP}} = 1 - \max_{c \in C} \sigma(\mathbf{z})_c
\end{equation}
where $\sigma$ represents the softmax function, $\mathbf{z}$ is the logit vector, and $\sigma(\mathbf{z})_c$ is the softmax probability for class $c$.

\textbf{Theory:} MSP uses the maximum softmax probability as a confidence measure. Higher maximum probabilities indicate confident ID predictions, while lower values correspond to uncertain or potentially OOD regions.

\subsection{Maximum Logit (MaxLogit)}
\textbf{Mathematical Formulation:}
\begin{equation}
U_{\text{MaxLogit}} = -\max_{c \in C} \mathbf{z}_c
\end{equation}
where $\mathbf{z}_c$ denotes the pre-softmax logit for class $c$, $C$ is the total number of classes, and $\max$ selects the highest logit value across all classes.

\textbf{Theory:} MaxLogit utilizes the highest raw logit value as a direct confidence measure, bypassing probability calibration. Lower maximum logits indicate higher prediction uncertainty, with the negative sign ensuring higher scores correspond to more uncertain or out-of-distribution samples. This method operates directly on pre-softmax activations, avoiding potential overconfidence issues in probability calibration.

\subsection{Energy Score}
\textbf{Mathematical Formulation:}
\begin{equation}
U_{\text{Energy}} = -\log\sum_{c=1}^{C} \exp(\mathbf{z}_c)
\end{equation}
where $\mathbf{z}_c$ is the logit for class $c$, $C$ is the total number of classes, and $\log$ denotes the natural logarithm.

\textbf{Theory:} Energy measures the overall activation magnitude across all classes. During inference, models assign lower energy to confident in-distribution predictions, creating natural separation for anomaly detection. Higher energy indicates uncertain or out-of-distribution samples.

\subsection{Entropy}
\textbf{Mathematical Formulation:}
\begin{equation}
U_{\text{Entropy}} = -\sum_{c=1}^{C} p_c \log p_c
\end{equation}
where $p_c = \sigma(\mathbf{z})_c$ is the softmax probability for class $c$, $C$ is the total number of classes, and $\log$ denotes the natural logarithm.

\textbf{Theory:} Entropy measures the information-theoretic uncertainty in the probability distribution. Maximum entropy occurs when all classes are equally likely, indicating high prediction uncertainty, while minimum entropy (zero) corresponds to completely confident predictions with one class having probability 1.

\subsection{Rejected by All (RbA)}
RbA~\cite{nayal2023rba} derives anomaly evidence from the absence of support across the model's semantic predictions. A point receives a high anomaly score when it is rejected by all known-class predictions, rather than relying only on the most confident class. We apply RbA to the native outputs of each segmentation architecture and orient all scores such that larger values indicate a higher likelihood of OOD.

\section{Societal Impact}

DriveOSS supports research on robust perception by exposing failures caused by unknown road objects and adverse environmental conditions. A standardized benchmark can help quantify these failures and guide safer perception systems. However, benchmark performance does not establish deployment readiness: the dataset has finite geographic, sensor, and anomaly coverage, and models may fail on unrepresented conditions. DriveOSS should therefore complement, rather than replace, broader safety validation. Privacy risks are mitigated through automatic anonymization and manual verification of released RGB images.

\section{Qualitative Results}
\label{sec:Visualization}

\begin{figure*}[h]
  \centering
  \includegraphics[width=\textwidth]{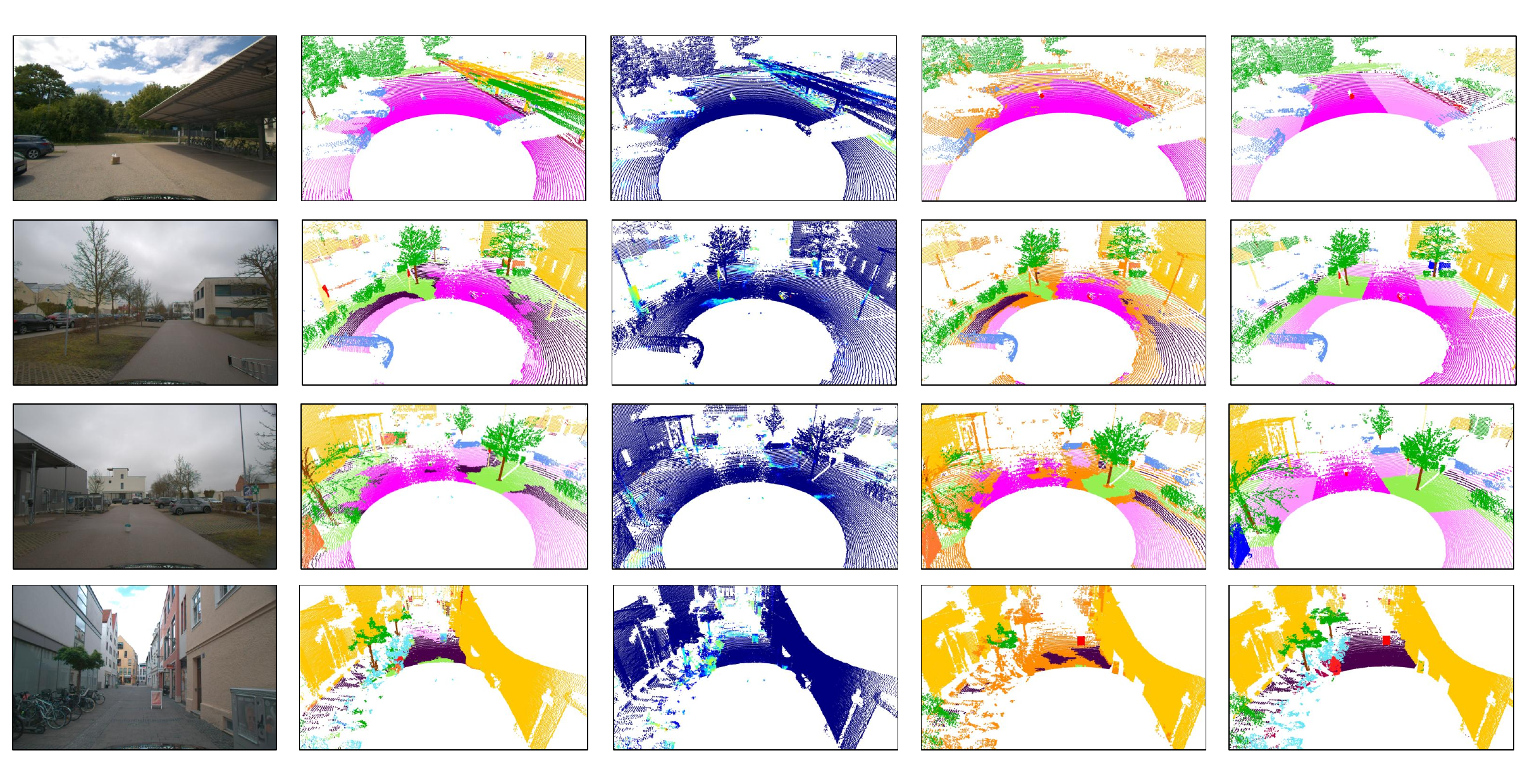}
  \caption{\textbf{Qualitative segmentation results.}
  From left to right: RGB image, closed-set semantic segmentation
  prediction, anomaly segmentation prediction, open-set semantic
  segmentation prediction, and open-set semantic segmentation ground truth.}
  \label{fig:results_vis}
\end{figure*}

Figure~\ref{fig:results_vis} qualitatively compares the three benchmark outputs and the final open-set ground truth. Closed-set segmentation assigns every point to an ID class, whereas anomaly segmentation highlights candidate OOD regions. Open-set semantic segmentation combines both outputs to preserve known-class predictions while assigning detected unknown points to \emph{outlier}.

\begin{figure*}[h]
  \centering
  \includegraphics[width=\textwidth]{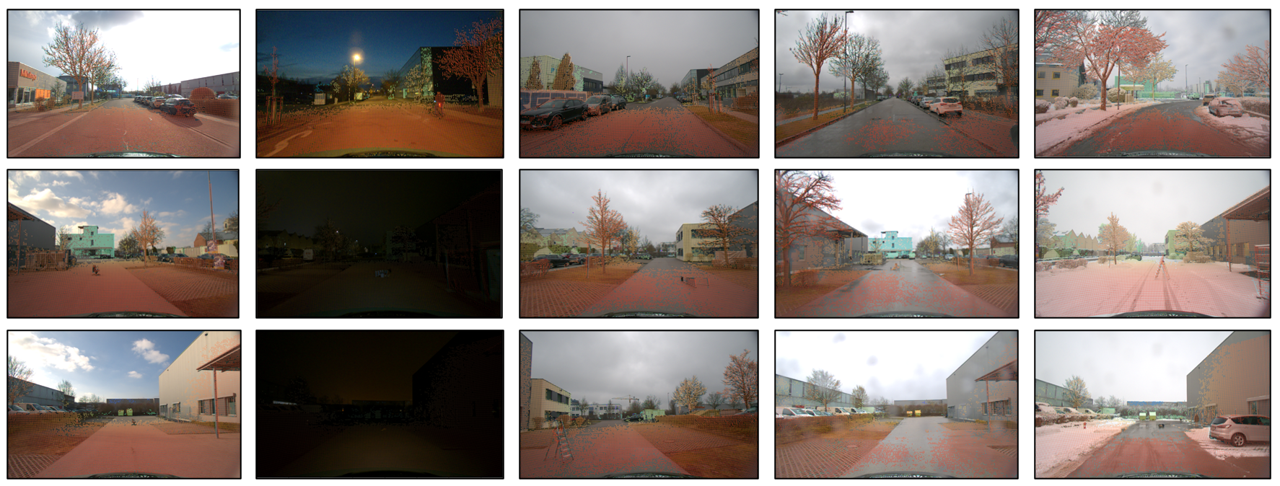}
  \caption{\textbf{Point cloud-to-image projection under different weather conditions.}
  From left to right, the columns show clear, night, cloudy, rainy, and snowy conditions.
  From top to bottom, the rows correspond to DriveOSS-ID, DriveOSS-OOD validation, and DriveOSS-OOD test scenes.
  The projected points show weather-dependent LiDAR distributions, with notably fewer road-surface reflections in rainy scenes.}
  \label{fig:projection_weather}
\end{figure*}

Figure~\ref{fig:projection_weather} shows representative samples from each environmental condition and data split, with LiDAR points projected onto the corresponding images using a uniform color. The projections illustrate condition-dependent variations in LiDAR returns and qualitatively verify the LiDAR--camera calibration. Rainy scenes exhibit visibly fewer road-surface returns, highlighting the sensing challenges faced by LiDAR and multimodal perception under adverse conditions.

\begin{figure*}[!h]
    \centering
    \includegraphics[width=\textwidth]{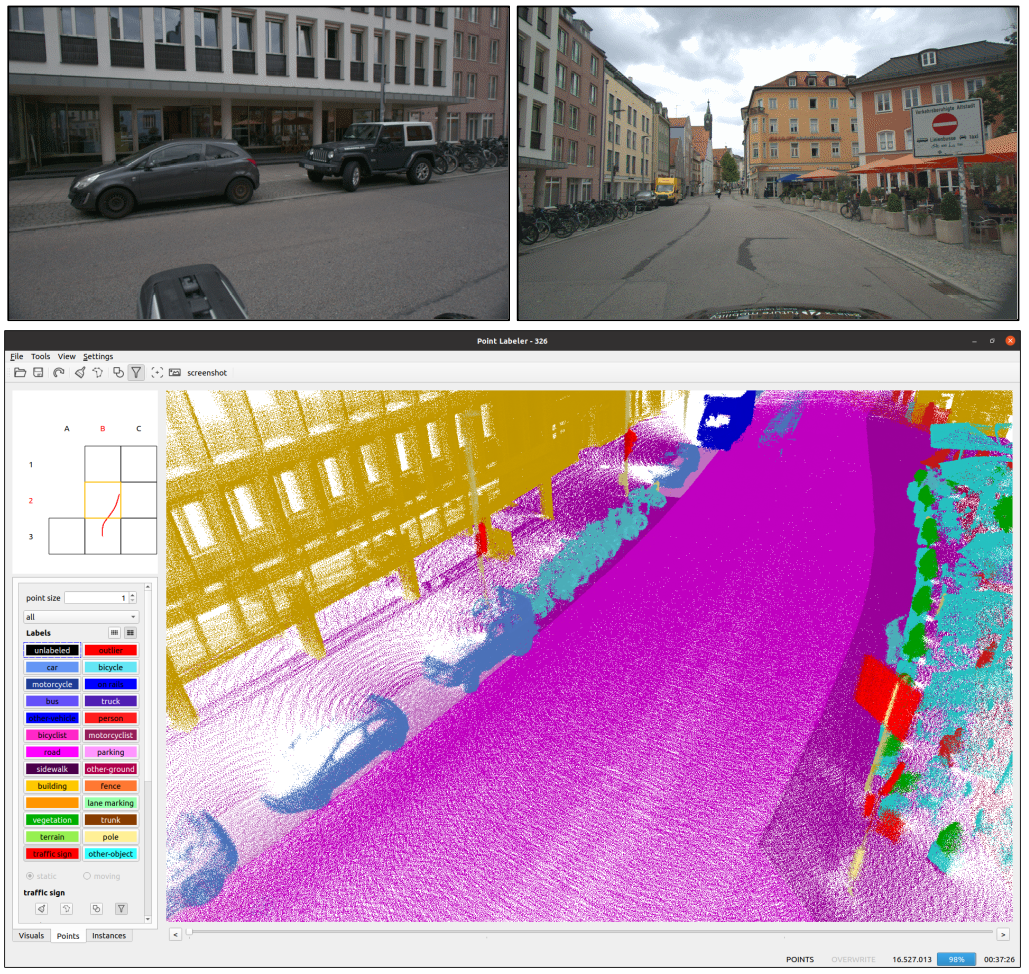}
    \caption{Snapshot of  the point cloud labeling interface, showing temporally registered 3D point clouds and RGB images.}
    \label{fig:labeler_ui}
\end{figure*}

\section{Annotation Workflow and Quality Control}
To construct accurate point-wise annotations across 12,661 frames while maintaining strict consistency, we employ a structured multi-stage annotation and quality-control (QC) workflow:
\begin{enumerate}
    \item \textbf{Coarse Pre-Labeling and Scan Registration:} We first apply a SemanticKITTI-pretrained Mask4Former model to generate coarse initial closed-set predictions. Consecutive LiDAR scans within each sequence are registered into a common coordinate frame using KISS-ICP to enable temporal context during labeling.
    \item \textbf{Inspection in the Point Labeler:} Registered sequences are loaded into the SemanticKITTI Point Cloud Labeler (illustrated in Fig.~\ref{fig:labeler_ui}). The six synchronized surround-view RGB images allow annotators to cross-reference 3D geometry with high-resolution visual appearance.
    \item \textbf{Ground and Boundary Refinement:} Annotators refine road, sidewalk, parking, and terrain boundaries by inspecting image features and applying the labeler's built-in \emph{remove ground} and \emph{remove plane} utilities to cleanly separate surface points from elevated obstacles.
    \item \textbf{Class-Wise Isolation Filtering:} Automatic pre-labeling frequently produces internal noise that is difficult to spot in a dense full-scene view (for instance, misclassifying interior points inside tree canopies as poles, buildings, or traffic signs rather than \emph{vegetation}). Annotators use the class-filter function to isolate individual classes one by one and manually correct falsely labeled points.
    \item \textbf{Manual OOD Annotation:} Because the pre-labeling network only predicts closed-set classes, annotators manually identify OOD objects and assign the \emph{outlier} label using synchronized camera views and 3D bounding-polygon tools.
    \item \textbf{Cross-Annotator Verification:} All annotators follow a strict, unified class definition and spatial-hazard policy. To improve inter-annotator consistency and mitigate individual bias, every annotated sequence undergoes independent manual cross-checking and verification by at least one additional annotator.
\end{enumerate}

Furthermore, we use manual refinement and cross-checking to mitigate potential bias introduced by model-generated pre-labels. The pretrained model does not include an explicit \emph{outlier} class, and all outlier labels in the final annotations are assigned manually. On DriveOSS-ID validation, M4F3D, which shares the architecture used for pre-labeling, achieves a closed-set mIoU of 34.98\%, compared with 58.35\% for LIDO and 54.05\% for LCPS. Although M4F3D performs worse than these alternatives, this comparison alone does not establish the absence of annotation bias.

\begin{figure*}[htbp]
    \centering
    \begin{tikzpicture}
        \node[anchor=south west, inner sep=0] (image) at (0,0) {\includegraphics[width=0.96\textwidth]{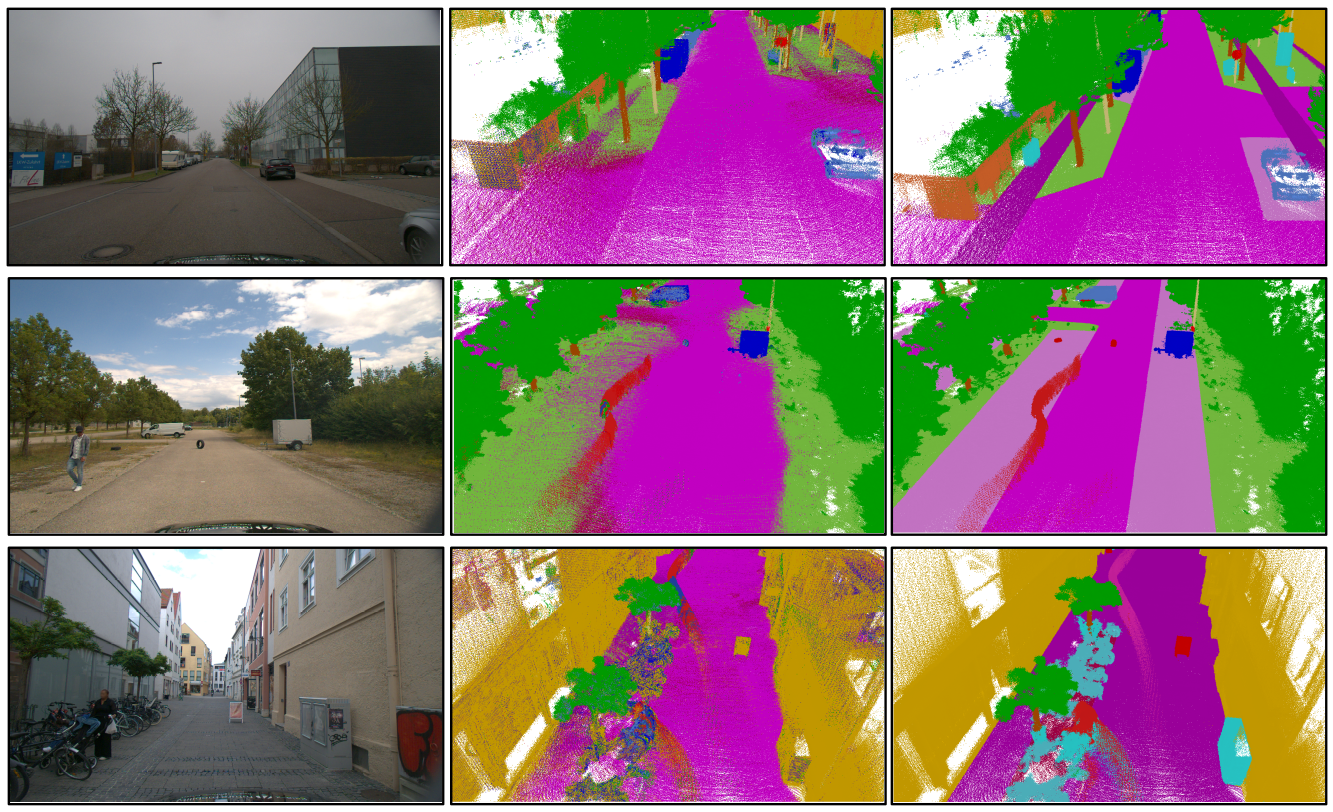}};

        \begin{scope}[x={(image.south east)},y={(image.north west)}]

            \node[align=center] at (0.167, 1.01) {RGB Image};
            \node[align=center] at (0.500, 1.01) {Automatic Prelabel};
            \node[align=center] at (0.833, 1.01) {Refined GT};

            \node[rotate=90, align=center] at (-0.01, 0.833) {DriveOSS ID};
            \node[rotate=90, align=center] at (-0.01, 0.500) {DriveOSS OOD Val};
            \node[rotate=90, align=center] at (-0.01, 0.167) {DriveOSS OOD Test};

        \end{scope}
    \end{tikzpicture}
    \caption{Visual comparison between automatic pre-labeling and final human-refined ground truth across DriveOSS subsets. Rows show representative samples from \textbf{ID}, \textbf{OOD Validation}, and \textbf{OOD Test}. Columns show: (1) RGB image, (2) raw Mask4Former pre-labels, and (3) final manual ground truth.}
    \label{fig:prelabel_vs_gt}
\end{figure*}

Figure~\ref{fig:prelabel_vs_gt} illustrates a side-by-side comparison of the RGB reference image, the raw automatic pre-labeling output, and the final human-refined ground truth across representative frames from DriveOSS-ID, DriveOSS-OOD validation, and DriveOSS-OOD test. The comparison illustrates corrections to initial pre-labels and the manual annotation of unknown road hazards in the displayed examples.

\end{document}